\pdfoutput=1
\documentclass[10pt]{article}
\let\keepaddcontentsline\addcontentsline
\usepackage[preprint]{tmlr}
\let\addcontentsline\keepaddcontentsline

\usepackage[english]{babel}

\usepackage{amsmath}
\usepackage{amssymb}
\usepackage{amsthm}
\usepackage{mathtools}
\usepackage{bm}
\usepackage{xcolor}

\DeclareMathOperator*{\argmax}{argmax}

\newtheorem{theorem}{Theorem}
\newtheorem{proposition}{Proposition}
\newtheorem{lemma}{Lemma}
\theoremstyle{definition}
\newtheorem{remark}{Remark}

\usepackage{graphicx}
\graphicspath{{figures/}}
\usepackage{booktabs}
\usepackage{subcaption}

\newcommand{\bleedgraphics}[2][1.25]{%
  \includegraphics[width=\linewidth]{#2}%
}
\usepackage[colorlinks=true, allcolors=blue]{hyperref}

\newcommand{\R}{\mathbb{R}}

\newcommand{\CC}{{\cal{C}}}
\newcommand{\CH}{{\cal{H}}}
\newcommand{\CX}{{\cal{X}}}
\newcommand{\cG}{{\cal{G}}}
\newcommand{\FE}{FELSTM}
\newcommand{\ME}{MELSTM}
\newcommand{\CL}{LSTM}

\makeatletter
\let\ifdeanonymized\if@accepted
\makeatother

\title{Motion Equivariant Convolutional LSTM:\\
       When Motion Becomes a Gauge Symmetry}

\author{\name Vahid Hosseinzadeh \\
        \addr Department of Mathematics \& Statistics\\
        South Dakota State University
        \AND
        \name T. Anderson Keller \\
        \addr The Kempner Institute at Harvard University\\
        Cusp AI
        \AND
        \name Michael Puthawala \\
        \addr Department of Mathematics \& Statistics\\
        South Dakota State University}

\begin{document}
\maketitle

\begin{abstract}
Objects move, cameras pan, and weather drifts with the wind: much of what changes in sequential data is motion. Motion changes where things are, not what they are, and is therefore a symmetry of the data. Yet standard recurrent networks are not equivariant to it: their memory stays put while the input moves on. Flow equivariant recurrent networks first brought this symmetry into sequence models, for motions of constant velocity, or flows, by enumerating a bank of velocities. We extend this idea in three directions: from simple recurrent networks to convolutional LSTMs, from flows to motions whose velocity changes in time, and from enumerating velocities to estimating them. Treating a sequence as a field over time, we show that motion becomes a gauge symmetry: a global symmetry made local in time. Built on this view, our motion equivariant convolutional LSTM (MELSTM) uses its velocity model as the gauge field and is provably equivariant to every motion. Across future-frame prediction, classification of digits visible only through their motion, and action recognition under camera shake, MELSTM outperforms a convolutional LSTM and its flow equivariant counterpart with the same number of parameters: it is more accurate, generalizes to longer horizons and to unseen motions, and carries a fraction of the flow equivariant model’s state.

\end{abstract}

\addtocontents{toc}{\protect\setcounter{tocdepth}{-1}}

\section{Introduction}
\label{sec:intro}

Symmetry is one of the most effective inductive biases in machine learning. A
model is \emph{equivariant} if transforming its input transforms its output in
the same way, so that its response to every transformed input follows from its
response to the original. Equivariance is usually built in by constraining the
weights. A linear map between fields is equivariant to translations exactly when
it is a convolution, whose weights are shared across positions
\citep{lecun1989backpropagation,kondor2018generalization}, and equivariant to
$G=\R^n\rtimes H$ when its kernel also satisfies a steerability constraint
\citep{cohen2017steerable,weiler2019general,cohen2019general}. The same principle
yields group convolutions \citep{cohen2016group}, convolutions on general Lie
groups \citep{finzi2020generalizing}, and gauge equivariant convolutions on
manifolds \citep{cohen2019gauge,weiler2026equivariant};
\citet{bronstein2021geometric} give a unified account. When the data have the
symmetry, equivariance provably improves generalization and sample efficiency
\citep{elesedy2021provably,bietti2021sample}. In almost all of this work,
however, the symmetry is static: when the input is a sequence, every frame is
transformed in the same way.

\begin{figure}
    \centering
    \includegraphics[width=1\linewidth]{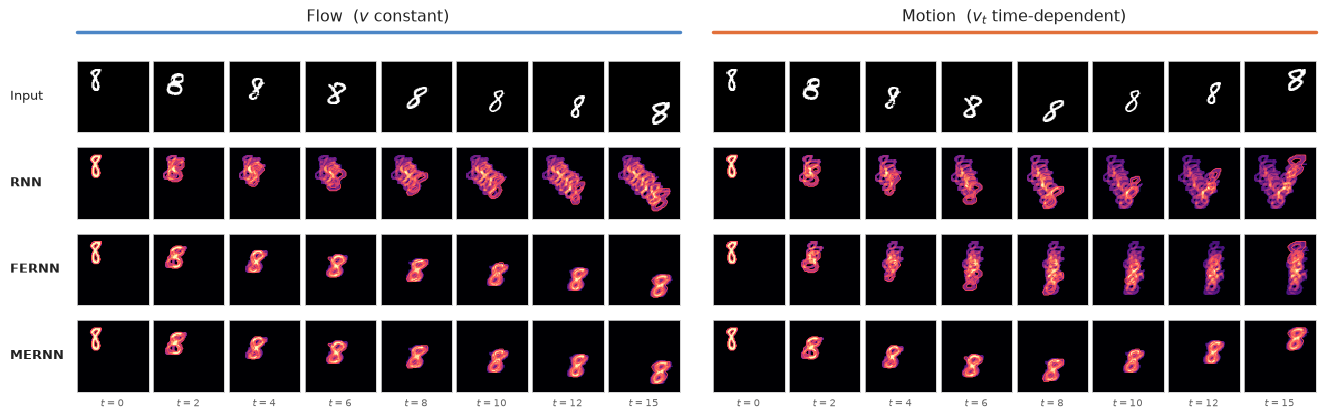}
\caption{\textbf{Hidden-state transport under RNN, Flow Equivariant RNN (FERNN), and Motion Equivariant RNN (MERNN) models.}
The first row shows the transformed input, $\phi\triangleright \CX_t$ in the flow column and $\psi \triangleright \CX_t$ in the motion column. Here an MNIST digit undergoes two simultaneous changes: a geometric transformation along the group orbit (the motion, which differs across columns) and an \emph{intrinsic appearance change} transverse to the group action (identical across columns).
The second row shows the hidden state of an untrained leaky accumulator, $h_{t+1} = \lambda\,h_t + \CX_t$ ($\lambda=0.92$), which can be viewed as a convolutional RNN,
$h_{t+1} = W \star h_t + U \star \CX_t$,
with kernels $W(x) = \lambda \delta(x)$ and $U(x) = \delta(x)$.
The third row shows FERNN's hidden state, which is lifted to velocity copies $g\in\{-3,\dots,3\}^2$ and evolves as
$h^g_{t+1}=\lambda\,g \cdot h^g_t+\CX_t$;
we show the copy of maximum energy, i.e., the copy with the largest $\|h_t^g\|_2^2$.
The fourth row shows MERNN's hidden state,
$h_{t+1} = \lambda T_{v_t}\,h_t + \CX_t$,
where $T_{v_t}$ is the transport operator and $v_t$ is the digit's velocity at time $t$.
Matched transport stacks evidence into a sharp digit at the true location, whereas mismatched transport leaves a decaying ghost trail along the residual motion.}
    \label{fig:motion_vs_flows}
\end{figure}

Sequential data transform in time: objects move and turn, cameras pan and shake,
and weather systems drift with the wind. The transformation that relates a
sequence to its transformed copy is then not one group element but a path
$t\mapsto\gamma_t$ in $G$. These paths form the group $\cG=G^{\R}$ of dynamical
symmetries (Section~\ref{sec:dynamical_symmetries}), which contains the
static, or global, symmetries as constant paths, the \emph{flows} as paths of
constant velocity, and the \emph{motions} as continuous paths whose velocity may
change in time. A recurrent model with $G$-equivariant layers, such as the
convolutional LSTM \citep[\CL{};][]{shi2015convolutionalLSTM} with steerable
kernels, is equivariant to the constant paths but not, in general, to the others:
it updates its memory in place rather than transporting it, so a moving input
leaves a decaying trail in the hidden state (Figure~\ref{fig:motion_vs_flows},
second row). One may ask whether flow or motion equivariance can also be obtained
by constraining the weights. On space-time, flows of translations are Galilean
boosts $(x,t)\mapsto(x+\nu t,t)$, whose orbits are entire time slices, so the
steerability constraint forces a space-time kernel to be constant in space at
every nonzero time lag; Lorentz boosts, in contrast, have hyperbolic orbits, and
Lorentz invariant kernels can vary in space (Remark~\ref{rem:spacetime}). For the
\CL{}, flow equivariance restricts the kernels to a set of measure zero
(Proposition~\ref{prop:noneq}). The weights alone cannot provide this symmetry:
the memory itself has to move with the input.

\citet{keller2025flowequivariantrecurrentneural} took the first step beyond
static symmetries with flow equivariant RNNs (FERNNs). FERNN lifts the hidden
state to a bank of copies, one for each velocity of a fixed lattice and each
moving with it, so an input moving with velocity $\nu$ is static in copy $\nu$
and accumulates there as if it did not move. Flows are, however, a narrow
class. They form a group only when $G$ is abelian
(Lemma~\ref{lemma-flows-subgroup}); when the velocity changes, the content of one
copy is handed over to another and the history accumulated in the first is lost
(Figure~\ref{fig:motion_vs_flows}, third row); and the bank must enumerate every
velocity the data may contain, which multiplies the size of the state and leaves
velocities outside the lattice unrepresented.

We take a different route: rather than enumerating velocities, we estimate them. A velocity model, trainable or parameter-free, reads the sequence and returns at
every step a Lie algebra
element $u_t\in\mathfrak g$, which generates a transporter $T_t\in G$ that
carries the hidden and cell states from one step to the next
(Figure~\ref{fig:motion_vs_flows}, fourth row). The resulting motion equivariant
\CL{} (\ME{}) is equivariant to all motions, provided the velocity model
transforms as $u_t\mapsto\psi_t u_t\psi_t^{-1}+\dot\psi_t\psi_t^{-1}$ when its
input is transformed by a motion $\psi$ (Theorem~\ref{thm:motion_equivariance}).
This is the transformation law of a gauge connection. In physics, making a global
symmetry local, so that the transformation may vary from point to point, requires
a connection and the parallel transport it defines
\citep{yang1954conservation,nakahara2003geometry}. Time-dependent transformations
are the global symmetries made local in time: a sequence is a section of a bundle
over the time axis, motions act on it as gauge transformations, the velocity
model is a connection, and the transporter is its parallel transport
(Appendix~\ref{app:gauge_theory}).
Gauge equivariant networks apply the same principle in space, to the choice of
frame on a manifold \citep{cohen2019gauge,weiler2026equivariant}; here the gauge freedom is the choice of frame at each instant.

So far, a single motion acts on the whole frame. Scenes, however, usually
consist of many independent motions: objects move on their own, against a
background that moves with the camera. The symmetry is then a \emph{local
motion}, which moves each component by its own motion, and a motion equivariant
model that respects superposition is also equivariant to local motions
(Section~\ref{sec:spatiotemporal}). \ME{} represents such scenes with a few
slots, each transported by its own velocity. With \emph{slot tracking}, each
slot estimates this velocity by matching its own hidden state against the
incoming frame, so the scene is segmented within the state and no correspondence
between frames has to be solved (Appendix~\ref{app:velocity_model}).
Our contributions are:
\begin{itemize}
    \item \textbf{Weight constraints are not enough.} With steerable kernels, the
    \CL{} is globally $G$-equivariant, yet it is flow equivariant only for a
    measure-zero set of local kernels. For space-time convolutions, Galilean
    steerability forces the kernel to be constant in space at every nonzero time
    lag (Proposition~\ref{prop:noneq}, Remark~\ref{rem:spacetime},
    Appendices~\ref{G-equivariance_of_ConvLSTM_model}--\ref{flow_equivariance_implies_constant_kernels_for_DenseLSTM}).
    \item \textbf{Flow equivariance by lifting.} Flows form a group only when $G$
    is abelian (Lemma~\ref{lemma-flows-subgroup}). We characterize the equivariant
    linear maps between transport fields and, for an abelian subgroup $V\le G$,
    build from them \FE{}, a flow equivariant \CL{}; the proof of its
    equivariance isolates the single step that requires constant velocity
    (Theorem~\ref{thm:flow_equivariance},
    Appendices~\ref{app:transport_convolution}--\ref{channel_based_FEConvLSTM_equivariance}).
    \item \textbf{Motion equivariance by transport.} A transporter is motion
    equivariant if and only if its velocity transforms as a gauge connection, and
    \ME{}, which transports its states with such a transporter, is equivariant to
    all motions, which act on the data as gauge transformations over the time axis
    (Theorem~\ref{thm:motion_equivariance},
    Appendices~\ref{velocity_model_equivariance}--\ref{app:gauge_theory}).
    \item \textbf{Local motions.} A motion equivariant model that respects
    superposition is also equivariant to local motions, which move the components
    of a scene independently (Section~\ref{sec:spatiotemporal}). \ME{} represents
    such scenes with a few transported slots, each estimating the velocity of its
    own content, which removes the correspondence problem of frame-to-frame motion
    estimation (Appendix~\ref{app:velocity_model}).
    \item \textbf{Experiments.} We compare \CL{}, \FE{} and \ME{} at equal
    parameter counts on three tasks with time-dependent motion: future-frame
    prediction on Moving MNIST \citep{srivastava2015unsupervised},
    classification of Motion-Only MNIST digits, which no single frame reveals
    because only the common fate of their pixels defines them
    \citep{upadhyay2026timeblindness}, and action
    recognition on KTH \citep{schuldt2004recognizing} under a shaking camera.
    Since \FE{} is flow but not motion equivariant, the comparison isolates what
    a changing velocity costs a flow equivariant model. \ME{} is the most
    accurate model on all three tasks and the only one whose performance does
    not degrade when the motion statistics change; on Moving MNIST, it trains
    about six times faster per epoch than \FE{} and extrapolates to $8.5$ times
    the trained horizon with $37$ times lower error
    (Section~\ref{sec:experiments}). \ifdeanonymized
    Code is available at \url{https://github.com/VahidHosseinzadeh/MELSTM}.
    \fi
\end{itemize}
Section~\ref{sec:discussion} discusses related work and future
directions.

\section{Equivariance; static vs dynamical symmetries}
\label{sec:dynamical_symmetries}

The data used in machine learning often exhibit structure and symmetry.
Incorporating these symmetries as inductive bias leads to equivariant models,
which are typically more efficient and generalize better.
Most existing work focuses on \emph{static symmetries}, where we have a time independent signal $X : M \to V$ which is a vector-valued function on the domain $M$. The space of these signals is denoted by $F$ which admits a group action of a symmetry group $G$. The equivariant model in this case is a function $\Phi$ that commutes with the group action:
\begin{align}
    \Phi[g \cdot X] = g \cdot \Phi[X], \qquad \forall g \in G, \; X \in F.
\end{align}
This static setup can be generalized to the case of time dependent signals or trajectories $\CX : \R \to F$,
which map time $t$ to a function $\CX_t \equiv \CX(t) \in F$. We denote the space of time dependent signals by ${\cal F} = \{ \CX : \R \to F \} \equiv F^{\R}$. The natural question is how to generalize the static symmetries to this dynamical setting.
The most general way is to extend the group $G$ to a group of time dependent transformations $\mathcal G := \{ \gamma : \R \to G \} \equiv G^\R$ where at each time $t$, we have a group element $\gamma_t \equiv \gamma(t) \in G$, and the group composition is defined pointwise ${(\gamma^{(1)} \cdot \gamma^{(2)})}_t := \gamma^{(1)}_t \gamma^{(2)}_t$.
Now the action of a time dependent transformation $\gamma \in \mathcal G$ on a time dependent signal $\CX \in \cal F$ is given by
\begin{align}
    {(\gamma \triangleright \CX)}_t := \gamma_t \cdot \CX_t, \qquad \forall t \in \R.
\end{align}
This defines a \emph{dynamical symmetry}, where transformations may vary arbitrarily over time.

\paragraph{Motions ${\cal G}_{mot}$.}
While $\mathcal G = G^\R$ allows arbitrary time dependence, many physically and
geometrically meaningful transformations are obtained by imposing regularity
conditions such as continuity or smoothness. A continuous map
$\psi : \R \to G$, with $G$ a topological group, is called a motion.
When $G$ is a Lie group, we assume a motion is differentiable and generated by a time-dependent Lie algebra valued function $v : \R \to \mathfrak g$, which can be interpreted as the \emph{velocity} of the motion. The evolution is governed by
$\frac{d}{dt} \psi_t = v_t \psi_t $ with initial condition $\psi_0 = e$, where $e$ is the identity element of $G$.
The motion $\psi$ generated by $v$ can be expressed as a time-ordered exponential, defined as the continuum limit of an ordered product of infinitesimal exponentials:
\begin{align}\label{motion_equation}
    \psi_t= \mathcal{T}\exp\left(\int_0^t v_s\, ds \right) =\lim_{N \to \infty} \left[ e^{v(\tau_N) \delta t} \cdots e^{v(\tau_1) \delta t} \right]
\end{align}
where $\delta t = \frac{t}{N}$ and ${\tau_i} = i \delta t$.
Note that the composition of two motions and the inverse of any motion are again motions. Therefore the set of motions is a subgroup of $\mathcal G$, denoted ${\mathcal G}_{mot} \le \mathcal G$. A proof appears in Appendix~\ref{structure_of_subgroups_of_G}.

\paragraph{Flows ${\cal G}_{flow}$.}
An interesting subset of motions are motions with constant velocity $v_t = \nu$ for all $t$. We call these motions \emph{flows} and denote them by $\phi$. In this case, equation~\eqref{motion_equation} reduces to $\phi_t = \exp(t \nu)$.
Equivalently, flows are motions $\phi:\R \to G$ that correspond to continuous \emph{group homomorphisms} from $\R$, viewed as the additive abelian group of time translations. The group homomorphism condition implies $\phi_{t+s} = \phi_t\phi_s$, which follows directly from equation~\eqref{motion_equation} as well. Note that flows form a subgroup of $\mathcal G$ if and only if $G$ is an abelian group. See Appendix~\ref{structure_of_subgroups_of_G} for proof.

\paragraph{Global symmetries ${\cal G}_{glob}$.}
The group $\cal G$ contains the subgroup of time-independent transformations, i.e., constant functions $\gamma(t) = g \in G$ for all $t$, which we call \emph{global transformations}. This subgroup is isomorphic to $G$, with elements denoted simply by $g$ where it acts uniformly across time via $(g \cdot \CX)_t = g \cdot \CX_t$.

\paragraph{Discrete vs Continuous time.}
In practice we often have a finite sample from the trajectory $\CX_t$ as  $\{\CX_t\}_{t=0}^{T}$. The definitions of symmetries and equivariance can be adapted to this setting by replacing $\R$ with the discrete set $\{0,1,\ldots,T\}$ and considering functions $g:\{0,1,\ldots,T\} \to G$. The same classification into motion, flow and global symmetries applies in the discrete case as well.
In the case of motions we have the recursive relation $\psi_t = \exp(v_t)\psi _{t-1}$ with $\psi_0 = e$ or equivalently $\psi_{t} = \prod_{s=1}^{t} \exp(v_s)$ for a sequence of velocities ${\{v_t\}}_{t=1}^{T}$, which is the discrete analogue of equation~\eqref{motion_equation}. In the case of flows we have $\phi_t = \exp(t\nu) ={(\phi_1)}^t$.

\section{Spatiotemporal data and Equivariant Sequential Models}
\label{sec:spatiotemporal}
\paragraph{Spatial data and convolution.} The formalism above is general and applies to any type of data and symmetries. However, here we focus on spatiotemporal data where the domain of signals $M$ is $\R^2$, the feature space $V$ is $\R^c$, and $F = {L}^2(\R^2; \R^c)$ is the space of square-integrable functions on $\R^2$ with values in $\R^c$. The group $G$ is typically spatial translations $(\R^2, +)$, $SE(2) = \R^2 \rtimes SO(2)$ (including rotations), or more generally $\R^2 \rtimes H$ where $H$ is a subgroup of $GL(2)$. We consider regular fields where the group action is defined as $(g \cdot X)(x) = X(g^{-1} \cdot x)$ for a $g \in G$. Note that this means we have a direct sum of $c$ number of scalar regular fields as the channels are not mixing under the group action.

Now, consider a linear map from a regular field $X \in F_{in} = {\cal L}^2(\R^2; \R^{c_{in}})$ to another regular field in $F_{out} = {L}^2(\R^2; \R^{c_{out}})$ via an integral operator ${\cal I}[X] = \int_{\R^2} W(x,y) X(y) dy$ with kernel $W: \R^2 \times \R^2 \to \R^{c_{out} \times c_{in}}$. For this map to be equivariant, it must be a convolution
\begin{align}
    [W \star X](x) = \int_{\R^2} W(x-y) X(y) dy
\end{align}
with a kernel satisfying the steerability constraint $W(h \cdot x) = \frac{1}{|{\det(h)}|}W(x)$ for all $h \in H$ and $x \in \R^2$. For instance, when $G = \R^2$, the equivariant map is standard convolution with no kernel constraints, while for $G = SE(2)$, it is convolution with a rotation invariant kernel $W(R \cdot x) = W(x)$ for all $R \in SO(2)$.

\paragraph{Spatiotemporal data and equivariant sequential models.}
We model spatiotemporal data as a trajectory $\CX : \R \to F$, where $F = {L}^2(\R^2; \R^c)$ is the space of regular fields. In discrete time, the temporal domain is replaced by $\{0,1,\ldots,T\}$ and the trajectory is represented by the sequence $(\CX_0,\CX_1,\ldots,\CX_T) \equiv \CX_{\le T}$
where each $\CX_t \in {L}^2(\R^2; \R^c)$ is a spatial field at time $t$.
Given the action of $G$ on $F$, we obtain an induced action of the trajectory group ${\cal G}$ on the trajectory space ${\cal F}$ defined by
\begin{align}
{(\gamma \triangleright \CX)}_t(x) = (\gamma_t \cdot \CX_t)(x) = \CX_t(\gamma_t^{-1} \cdot x)
\end{align}
for $\gamma \in {\cal G}$ and $\CX \in {\cal F}$.

A sequential model is a map $\Phi$ that takes as input the history $\CX_{\le t}$ and outputs a regular field $\Phi[\CX_{\le t}] \in F$. Such a representation may then be used for forecasting, sequence classification, or other downstream tasks. We say that the model is \emph{motion equivariant} if
\begin{align}
\Phi[\psi \triangleright \CX_{\le t}] = \psi \triangleright \Phi[\CX_{\le t}]
\qquad
\forall \psi \in {\cal G}_{mot}, \; \forall \CX \in{\cal F}.
\end{align}
Similarly, replacing $\psi \in {\cal G}_{mot}$ with $\phi \in {\cal G}_{flow}$ or $g \in {\cal G}_{glob}$ yields the notions of flow equivariance and global $G$-equivariance, respectively.

\paragraph{Local motion equivariance.} The transformations above act on the whole signal at
once. In practice, however, the data of interest is often a superposition of several objects,
each carrying its own motion. Given a decomposition $\CX_t=\sum_{\alpha=1}^K \CX^\alpha_t$ and a
tuple $\psi=(\psi^1,\dots,\psi^K)\in\cG_{mot}^K$, a \emph{local motion} acts by
$(\psi \triangleright \CX)_t \;:=\; \sum_\alpha (\psi^\alpha \triangleright \CX^\alpha)_t$,
and we speak of a \emph{local flow} when every $\psi^\alpha$ is a flow. Now suppose that the
model respects the superposition property, in the sense that
$\Phi[\sum_\alpha \CX_{\le t}^\alpha] = \sum_\alpha \Phi[\CX_{\le t}^\alpha]$. Then, if the
model is motion equivariant in the sense above, it is automatically equivariant under local
motions as well:
\begin{align}
    \Phi\Big[\sum_\alpha \psi^\alpha \triangleright \CX^\alpha_{\le t}\Big]
    =\sum_\alpha \Phi\Big[\psi^\alpha \triangleright \CX^\alpha_{\le t}\Big]
    =\sum_\alpha \psi^\alpha \triangleright \Phi\big[\CX^\alpha_{\le t}\big]
    \qquad \forall\, \psi \in {\cal G}_{mot}^K ,\; \forall \CX \in{\cal F},
\end{align}
where the first equality uses the superposition property and the second uses motion
equivariance.

\section{Motion Equivariant Convolutional LSTM}
\label{sec:models}
\paragraph{\CL{}.}
The Convolutional LSTM (\CL{}) \citep{shi2015convolutionalLSTM} is a first-order Markovian recurrent model with two state variables, the cell state $\{\CC_t\}$ and hidden state $\{\CH_t\}$, evolving according to
\begin{align}\label{ConvLSTM_equations_1}
z_t &= a_z\big(W_z \star {\cal H}_t + U_z \star {\cal X}_t\big) \qquad \forall z \in \{i,f,o,c\} \\
{\cal C}_{t+1} &= {\cal C}_t \odot f_t + i_t \odot c_t \\
{\cal H}_{t+1} &= o_t \odot \tanh({\cal C}_{t+1}) \label{ConvLSTM_equations_6}
\end{align}
where $\odot$ denotes element-wise multiplication, $\{i_t,f_t,o_t,c_t\}$ are the input, forget, output, and cell gates, and the nonlinearity $a_z$ is $\sigma$ for $z \in \{i,f,o\}$ and $\tanh$ for $z = c$.

The \CL{} is globally $G$-equivariant when the convolutional kernels $W_z$ and $U_z$ are chosen to satisfy the appropriate steerability constraints for the group $G$. See Appendix~\ref{G-equivariance_of_ConvLSTM_model} for detailed statements and proof.

The \CL{}, however, is not flow equivariant for generic kernels
(Proposition~\ref{prop:noneq}, Appendix~\ref{flow_equivariance_implies_constant_kernels_for_DenseLSTM}). The obstruction is structural: \CL{} updates its memory in place rather than transporting it with the flow. As illustrated for a simple RNN in the second row of Figure~\ref{fig:motion_vs_flows}, this leaves a decaying trail, or left-behind memory, instead of transporting the accumulated state with moving features. The appendix shows that eliminating this residual memory requires an exact cancellation among the local convolution kernels, satisfied only by a measure-zero set of parameters. Transporting the hidden and cell states with the flow removes this obstruction, motivating the architectures introduced next.

\paragraph{Channel based flow equivariant \CL{} (\FE{}).}
In this section, following the Flow Equivariant RNN (FERNN) introduced by
\citet{keller2025flowequivariantrecurrentneural}, we introduce the channel based
\FE{}, a generalization of the \CL{} architecture that achieves flow
equivariance by lifting the regular fields on $M$ to the larger space $M \times G$.
In particular the gate variables transform under $\cG$ as
\begin{align}
{(\gamma \triangleright z)}_t(x, g)= \gamma_{t} \cdot z_t\big(x, \,\gamma_t^{-1} g\,\gamma_{t-1} \big) &=  z_t\big( \gamma_{t}^{-1}\cdot x,\,\gamma_t^{-1} g\,\gamma_{t-1} \big)
\end{align}
and the state variables $\mathcal S \in \{\CH, \CC\}$ transform as\footnote{Note that the state variable at time $t$ is a function of the input sequence up to time $t-1$ so it transforms according to $\gamma_{t-1}$ instead of $\gamma_t$.}
\begin{align}
{(\gamma \triangleright \mathcal S)}_t(x, g)= \gamma_{t-1} \cdot\mathcal S_t\big(x, \,\gamma_t^{-1} g\,\gamma_{t-1} \big) &=  \mathcal S_t\big( \gamma_{t-1}^{-1}\cdot x,\,\gamma_t^{-1} g\,\gamma_{t-1} \big).
\end{align}
We refer to the transformation $g \mapsto \gamma_t^{-1} g\,\gamma_{t-1}$
as \emph{transport conjugation}, and fields on $M \times G$ whose transformation law
involves this action on the velocity coordinate are called \emph{transport
fields}.\footnote{This terminology is inspired by the parallel transport concept in
gauge theories which is the theoretical basis of our transporter based model in the
next section.}
As we saw earlier, $\cG_{flow}$ is a group (subgroup of $\cG$) if and only if $G$ is abelian (see Lemma~\ref{lemma-flows-subgroup}). So for general non-abelian groups, we restrict to an abelian subgroup $V \le G$ and build the flow equivariant model for ${\cal V}_{flow} \le \cal G$ as the symmetry group. Therefore we lift to transport fields on $M \times V$ and modify update equations as follows
\begin{align}\label{channel_based_FEConvLSTM_equations-1}
z_t(x, \mathsf{v}) &= a_z\Big( \mathsf{v} \cdot [W_z \star_F {\cal H}_t](x, \mathsf{v}) + [U_z \, \hat{\star}\, {\cal X}_t](x,  \mathsf{v}) \Big) \qquad \forall z \in \{i,f,o,c\} \\
{\cal C}_{t+1}(x, \mathsf{v}) &= [\mathsf{v} \cdot {\cal C}_t(x, \mathsf{v})]  \odot f_{t}(x, \mathsf{v})  + i_{t}(x, \mathsf{v}) \odot c_{t}(x, \mathsf{v}) \\
{\cal H}_{t+1}(x, \mathsf{v}) &= o_{t}(x, \mathsf{v}) \odot \tanh({\cal C}_{t+1}(x, \mathsf{v}))\label{channel_based_FEConvLSTM_equations-3}
\end{align}
for all $z \in \{i,f,o,c\}$ and $\mathsf{v} \in V$.
Here $\hat{\star}$ is the flow-lifting convolution operator which is an equivariant linear map that lifts the input signal $\CX_t$ to the transport field space. It turns out that this map is simply given by copying the steerable convolution to all the channels indexed by $\mathsf{v}$ as follows $[U \, \hat{\star}\, \CX_t](x,\mathsf{v}) =[U  \!\star \!\CX_t](x)$.
Also $\star_F$ is the flow convolution  which is an equivariant linear map between transport fields and it is defined as follows
\begin{align}\label{flowConvolution}
    [W \star_F \CH_t](x,\mathsf{v}) = \int_{\cal M} \int_V W(x-y, \mathsf{v} - \mathsf{u}) \CH_t(y,\mathsf{u}) d\mathsf{u} dy
\end{align}
where the kernel $W$ must additionally satisfy a steerability constraint in its spatial argument.
See Appendix~\ref{app:transport_convolution} for the details of the proof of equivariance.

\begin{theorem}[\FE{} is flow equivariant]\label{thm:flow_equivariance}
Consider the channel-based \FE{} defined by equations~\eqref{channel_based_FEConvLSTM_equations-1}--\eqref{channel_based_FEConvLSTM_equations-3}, with invariant initial state variables satisfying $\phi \triangleright \CH_0 = \CH_0$ and $\phi \triangleright \CC_0 = \CC_0$ for all $\phi \in {\cal V}_{flow}$. Then the model is flow equivariant, i.e., for any flow $\phi \in {\cal V}_{flow}$ and any trajectory $\CX \in {\cal F}$, we have
\begin{align}
        \CH_{t}[\phi \triangleright \CX_{\le t}] = \phi \triangleright \CH_{t}[\CX_{\le t}] \qquad \forall t.
\end{align}
\end{theorem}
See Appendix~\ref{channel_based_FEConvLSTM_equivariance} for the proof.

\paragraph{Transporter based motion equivariant \CL{} (\ME{}).}
The channel-based \FE{} achieves equivariance under flows by lifting regular fields on $M$ to transport fields on $M \times G$, which can be interpreted as introducing a bank of moving frames indexed by the group $G$. However, this construction is equivariant only under flows ${\cal V}_{flow}$ of an abelian subgroup $V \le G$, not under general time-dependent motions $\cG_{mot}$; in practice, $V$ is even truncated to a finite grid of velocities.
Moreover, it increases the dimensionality of the hidden representation by introducing an additional group dimension. Here, we take a different approach. Instead of lifting the representation space, we explicitly transport the hidden states between consecutive time steps using an estimated group-valued transformation. This leads to a transporter-based architecture that achieves equivariance under general motions while keeping the fields regular and defined only on the original spatial domain $M$.

This estimated transformation is the \emph{transporter} $T_{t_2,t_1} \in G$, which describes how the signal transforms between times $t_1$ and $t_2$. It is generated by a Lie algebra valued velocity $u_t \in \mathfrak g$ through the ODE $\frac{d}{dt} T_{t,t_0} = u_t T_{t,t_0}$ with $T_{t_0,t_0} = e$, so that, like the motion $\psi_t$ generated by $v_t$ in \eqref{motion_equation}, it is a time-ordered exponential:
\begin{align}
T_{t_2,t_1} =\mathcal{T}\exp\left(
\int_{t_1}^{t_2} u_s ds \right)= \lim_{N \to \infty} \left[ e^{u(\tau_N) \delta t} \cdots e^{u(\tau_1) \delta t} \right]
\end{align}
where $\delta t = \frac{t_2-t_1}{N}$ and ${\tau_i} = t_1 + i \delta t$.
We also need the transporter to satisfy the equivariance relation as
\begin{align}\label{transporter_equivariance_relation}
T_{t_2,t_1}[\psi \triangleright \CX_{\le t_2}] = \psi_{t_2} \,T_{t_2,t_1}[\CX_{\le t_2}] \,\psi_{t_1}^{-1}
\end{align}
for every motion $\psi \in {\cal G}_{mot}$. In practice, we might model the instantaneous velocity $u_t$ instead of the transporter $T_{t_2,t_1}$ directly. Using the ODE governing the transporter, we can show that this equivariance relation is satisfied if and only if the velocity model $u_t$ satisfies the following equivariance relation (see Appendix~\ref{velocity_model_equivariance} for proof):
\begin{align}\label{eq:velocity_law}
u_t[\psi \triangleright \CX_{\le t}] = \psi_t u_t[\CX_{\le t}] \psi_t^{-1} + \dot{\psi}_t \psi_t^{-1}
\end{align}
for every motion $\psi \in {\cal G}_{mot}$. This is precisely how a gauge connection transforms. Viewing a sequence as a section of a bundle over the time axis, the spatial coordinates of each $\CX_t$ may be chosen independently at every instant: motions change these coordinates as gauge transformations, the velocity model $u_t$ is the connection that relates coordinates at neighboring times, and the transporter is its parallel transport (Appendix~\ref{app:gauge_theory}).

In the discrete-time setting, we denote $T_t \equiv T_{t,t-1}$ as the transporter between two consecutive time steps. In this case, we simply have $T_t = \exp(u_t)$.
Using the transporter, the \ME{} is defined by the following equations
\begin{align}\label{MEConvLSTM_equations-1}
z_t(x) &= a_z\Big( T_t\cdot [W_z \star {\cal H}_t](x) + [U_z \star {\cal X}_t](x) \Big) \qquad \forall z \in \{i,f,o,c\} \\
{\cal C}_{t+1}(x) &= [T_t\cdot {\cal C}_t(x)] \odot f_t(x)  + i_t(x)\odot c_t(x) \\
{\cal H}_{t+1}(x) &= o_{t}(x) \odot \tanh({\cal C}_{t+1}(x))\label{MEConvLSTM_equations-3}
\end{align}

\begin{theorem}[\ME{} is motion equivariant]\label{thm:motion_equivariance}
Consider the transporter-based \ME{} defined by equations~\eqref{MEConvLSTM_equations-1}--\eqref{MEConvLSTM_equations-3}, with invariant initial state variables satisfying $\psi \triangleright \CH_0 = \CH_0$ and $\psi \triangleright \CC_0 = \CC_0$ for all $\psi \in {\cal G}_{mot}$. Then the model is motion equivariant, i.e., for any motion $\psi \in {\cal G}_{mot}$ and any trajectory $\CX \in {\cal F}$, we have
\begin{align}
        \CH_{t}[\psi \triangleright \CX_{\le t}] = \psi \triangleright \CH_{t}[\CX_{\le t}] \qquad \forall t.
\end{align}

\end{theorem}
\begin{proof}[Proof sketch.] The proof follows the same inductive structure as the flow equivariance result for \FE{}. The base case is immediate from the invariance of the initial state variables. Note that the transporter equivariance relation~\eqref{transporter_equivariance_relation} in the discrete-time setting reduces to $T_t[\psi \triangleright \CX_{\le t}] = \psi_t\, T_t[\CX_{\le t}]\, \psi_{t-1}^{-1}$. Assuming that the state variables are motion equivariant at time $t$,
we first show that the gate variables remain motion equivariant up to time $t$. The key step is the cancellation of the $\psi_{t-1}$ factor between the transporter and the hidden state in the gate equations. We then apply the same argument to the cell-state and hidden-state updates to conclude motion equivariance at time $t+1$. See Appendix~\ref{transporter_based_FEConvLSTM_equivariance} for the complete proof.
\end{proof}

\section{Experiments}
\label{sec:experiments}

We evaluate the proposed motion equivariant \CL{} (\ME{}) against the
flow equivariant \CL{} (\FE{}) and a standard \CL{} baseline on
prediction and classification under time-dependent motion. The three models
share an architecture, a parameter count, and a training protocol, and differ
only in how the hidden state is transported between timesteps, so any
difference in performance is attributable to that mechanism alone. Our aim is
not only to show that equivariance helps, which is established for flows
\citep{keller2025flowequivariantrecurrentneural}, but to ask what form it should
take when the motion itself varies in time: whether it is better to
\emph{enumerate} a lattice of candidate velocities or to \emph{estimate} a small
number of them.

\subsection{Moving MNIST}
\label{sec:mm}

Both experiments in this section use the same synthetic motion and the same
three encoders. They differ only in how the frames are rendered and in what the
final hidden state is used for.

\paragraph{Time-dependent motion.}
MNIST digits move on a $36\times36$ canvas with periodic boundary conditions.
Velocities are discrete, drawn from the integer lattice $V_2$ where
$V_n=\{v\in\mathbb Z^2\mid\|v\|_\infty\le n\}$, so that translation by any
$v\in V_2$ acts exactly on the periodic grid without interpolation. Motions are
piecewise constant: a velocity is held for a random $3$--$6$ frames and then
resampled. Every moving component $\alpha$ carries its own velocity schedule
$v^{(\alpha)}_t\in V_2$ and hence its own motion $\psi^{(\alpha)}$, and the models
read a context of $T=15$ frames.

\paragraph{Models.}
The three models share one recurrent cell and differ only in how its hidden
state is transported: not at all (\CL{}), by enumeration over $|V_2|=25$ velocity
channels (\FE{}), or by $K=2$ slots warped by velocities $u_t[\CX_{\le t}]$
estimated online by parameter-free Fourier phase correlation (\ME{};
Appendix~\ref{app:velocity_model}). The velocity dimension is max-pooled and
read by a task head $g_\theta$: a decoder that produces each predicted frame,
$\hat\CX_t=g_\theta\bigl(\max_v h_t(v)\bigr)$ (Section~\ref{sec:mm_pred}), or a
classifier applied once to the final state, $\hat y=g_\theta\bigl(\max_v
h_T(v)\bigr)$ (Section~\ref{sec:mm_cls}). Weights are shared across channels and
slots, so for each task the three models have identical parameter counts, and
they are trained identically.

\subsubsection{Future-Frame Prediction}
\label{sec:mm_pred}

\begin{figure}[t]
\centering
\begin{minipage}[c]{0.715\linewidth}
  \centering
  \includegraphics[width=\linewidth]{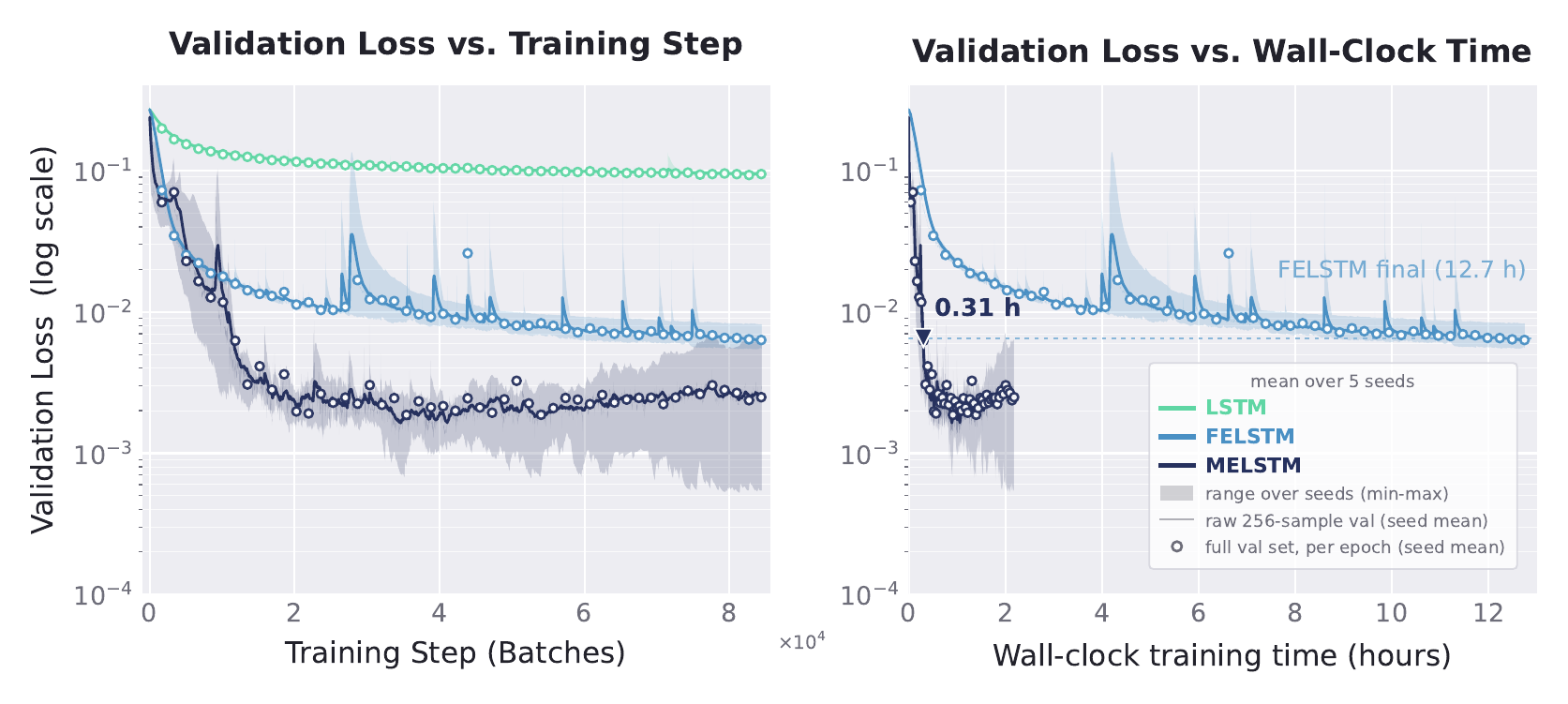}
\end{minipage}\hfill
\begin{minipage}[c]{0.285\linewidth}
  \centering
  \footnotesize
  \setlength{\tabcolsep}{1.5pt}
  \renewcommand{\arraystretch}{1.35}
  \begin{tabular}{lcc}
  \toprule
  Model & \begin{tabular}{@{}c@{}}MSE $\downarrow$\\[-1pt]{\tiny test}\\[-3pt]{\tiny $t=100$}\end{tabular}
        & \begin{tabular}{@{}c@{}}s /\\epoch\end{tabular} \\
  \midrule
  LSTM   & \begin{tabular}{@{}c@{}}$3.26_{\pm .05}$e$-$2\\[-2pt]$2.01_{\pm 1.34}$e$-$1\end{tabular} & $77_{\pm 1}$ \\
  \addlinespace[2pt]
  FELSTM & \begin{tabular}{@{}c@{}}$1.54_{\pm .42}$e$-$3\\[-2pt]$8.83_{\pm 6.58}$e$-$2\end{tabular} & $918_{\pm 9}$ \\
  \addlinespace[2pt]
  MELSTM & \begin{tabular}{@{}c@{}}$\bm{2.87_{\pm 2.20}}$\textbf{e}$\bm{-}$\textbf{4}\\[-2pt]$\bm{1.16_{\pm 1.05}}$\textbf{e}$\bm{-}$\textbf{3}\end{tabular} & $157_{\pm 5}$ \\
  \bottomrule
  \end{tabular}
\end{minipage}
\caption{\textbf{Training efficiency and test error.} \textbf{Right}: test MSE over the
trained horizon and at $t=100$; mean $\pm$ s.d.\ over five seeds.}
\label{fig:trainingeff}
\end{figure}

\begin{figure}[!b]
    \centering
    \includegraphics[width=1\linewidth]{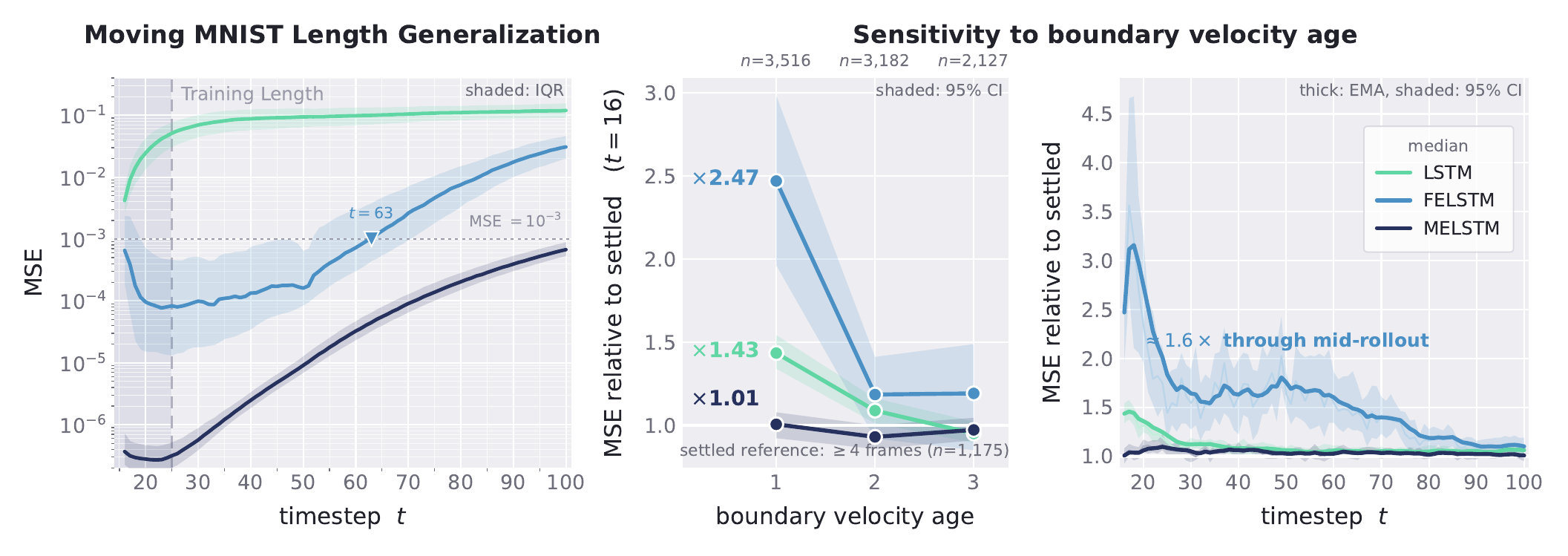}
\caption{\textbf{Rollout stability and context-boundary sensitivity.} \textbf{Left}:
median rollout MSE. \textbf{Middle}: MSE at $t=16$ of sequences whose last
velocity change came $1$--$3$ frames before the context boundary (boundary
velocity age), relative to settled sequences (age $\ge 4$). \textbf{Right}: the age-$1$
group over the rollout; the \FE{} penalty persists through mid-rollout before
gradually decaying.}
    \label{fig:len_gen_boundary_age}
\end{figure}

\begin{figure}
    \centering
    \includegraphics[width=1\linewidth]{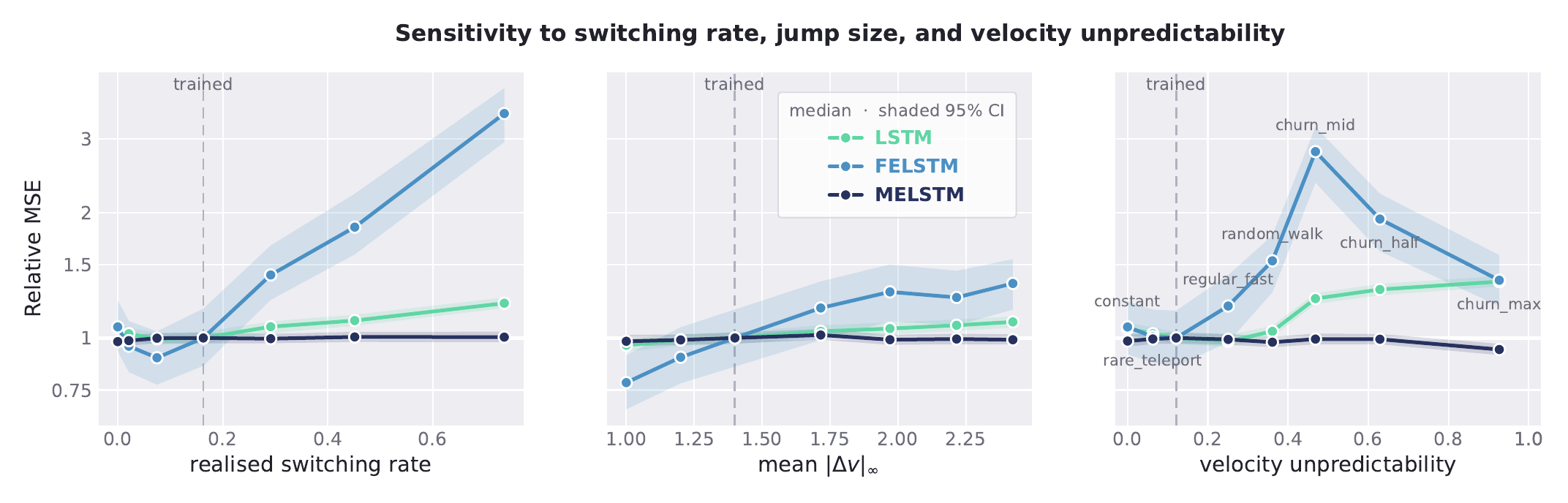}
\caption{\textbf{Motion generalization sensitivity.} Each model normalized by its own training error ($1.0$ = no degradation).}
\label{fig:motion_generalization}
\end{figure}

\paragraph{Data and task.}
Two digits move independently, so a sequence is a sum of two independently
transported components, $\psi^{(1)}\triangleright\CX^{(1)}+\psi^{(2)}\triangleright\CX^{(2)}$.
Every sequence has two phases: during the first $T = 15$ steps, the context,
each digit moves under its own time-dependent velocity schedule, and at the
context boundary the velocities of local motions are frozen, $v_s = v_T$ for all
$s > T$. Models are trained autoregressively on this structure: given the $15$
context steps of $\CX$, predict the following $10$, over which each local motion
reduces to a constant flow. \FE{} is flow-equivariant by construction, so
its task is to have encoded, by the end of the context, a good representation of
each digit in the velocity channel that matches the digit's final velocity.
\ME{} must encode each digit in one of its slots \emph{and} recover the
digit's final velocity, which it then uses to transport that slot through the
frozen phase. Parameters are optimized to minimize MSE\,$+$\,L1 error against
the ground truth over $t \in [16,25]$ for $50$ epochs, identical for every
model. Figure~\ref{fig:trainingeff} shows
the validation loss of the three models: \ME{} outperforms both \FE{}
and the standard \CL{}, demonstrating the benefits of motion equivariance in
handling time-dependent motions.

\paragraph{Slot tracking.}
For prediction, \ME{} takes
$u_t[\CX_{\le t}] = u_t\big[\CH_t[\CX_{<t}],\,\CX_t\big]$: each slot correlates
its own hidden state with the current frame. Since a slot carries a single
coherently moving component, the local problem of a velocity field over a
multi-object frame reduces to $K$ global displacements, one per slot, each given
in closed form by phase correlation. This also solves the correspondence (or
assignment) problem. Correlating two consecutive frames yields $K$ velocities per
step, but as an unordered set that does not say which velocity at $t$ continues
which at $t-1$; here each estimate is bound to the slot that requested it. We
call this \emph{slot tracking} (Appendix~\ref{app:velocity_model}). Note that early in training, slot tracking can also be replaced by frame-pair velocities, withdrawn gradually under a stochastic or curriculum schedule (Appendix~\ref{app:pred_xtrain}).

\paragraph{Length generalization.}
Figure~\ref{fig:len_gen_boundary_age} shows the prediction error when the frozen
phase is extended from $10$ to $85$ steps: \ME{} is lowest at every timestep, by $37\times$ over \FE{}
on average and on $99.7\%$ of individual predicted frames, with all models scored
on the same $10{,}000$ sequences. Stratifying by how recently a digit's
velocity last changed before the context ended locates where that advantage
comes from (Figure~\ref{fig:len_gen_boundary_age}): a change one frame before the
boundary costs \FE{} roughly $2.5\times$ its error at the first predicted
frame, above the baseline's own sensitivity to the same event, while \ME{}
pays nothing measurable. The enumerating model is therefore the most sensitive of
the three to a recent change in motion and the estimating model the least, even
though \ME{} is the only one that must recover the velocity and so the one
that would be expected to suffer most from a single frame of new evidence.

\paragraph{Motion generalization.}
We evaluate each model as trained on benchmarks whose motion statistics differ from the training distribution, scoring MSE over the same trained horizon and dividing each model by its own error at the training configuration; Figure~\ref{fig:motion_generalization} thus shows sensitivity, not accuracy. Two compensated sweeps vary switching rate and jump size independently. \FE{} keeps 25 hidden-state copies, one per possible velocity; when the velocity changes, it must read from a different copy and abandon what had accumulated. The penalty is roughly the product of how disruptive the velocity changes are,
which grows with both how often and how far the velocity jumps, and how much the
abandoned copy was worth. The rate sweep raises the first factor directly and
costs \FE{} $3.45\times$; larger jumps raise it less and cost $1.35\times$. \ME{} transports its two slots by an estimated velocity parameter, so a velocity change moves the same representation differently rather than transferring between velocity-indexed copies; it stays flat to within $1\%$ on both axes. \CL{} has no velocity-indexed state to disturb and tracks only motion difficulty. Seven fixed regimes probe combinations of rate and jump size beyond the swept axes. \CL{}'s rise measures intrinsic difficulty, while the ratio of \FE{} to \CL{} isolates architecture-specific cost. Figure~\ref{fig:motion_generalization} also orders these regimes by velocity unpredictability $\rho$, the error of
using the previous velocity relative to assuming no motion. $\rho$ measures the
disruption and $1-\rho$ the value of the abandoned history, so the penalty should
follow $\rho(1-\rho)$ (Appendix~\ref{app:length_motion_generalization}); \FE{}'s
penalty indeed peaks near $\rho=0.5$.

\subsubsection{Motion-Only MNIST Classification}
\label{sec:mm_cls}

\begin{figure}[!b]
\centering
\begin{minipage}[c]{0.725\linewidth}
  \centering
  \includegraphics[width=\linewidth]{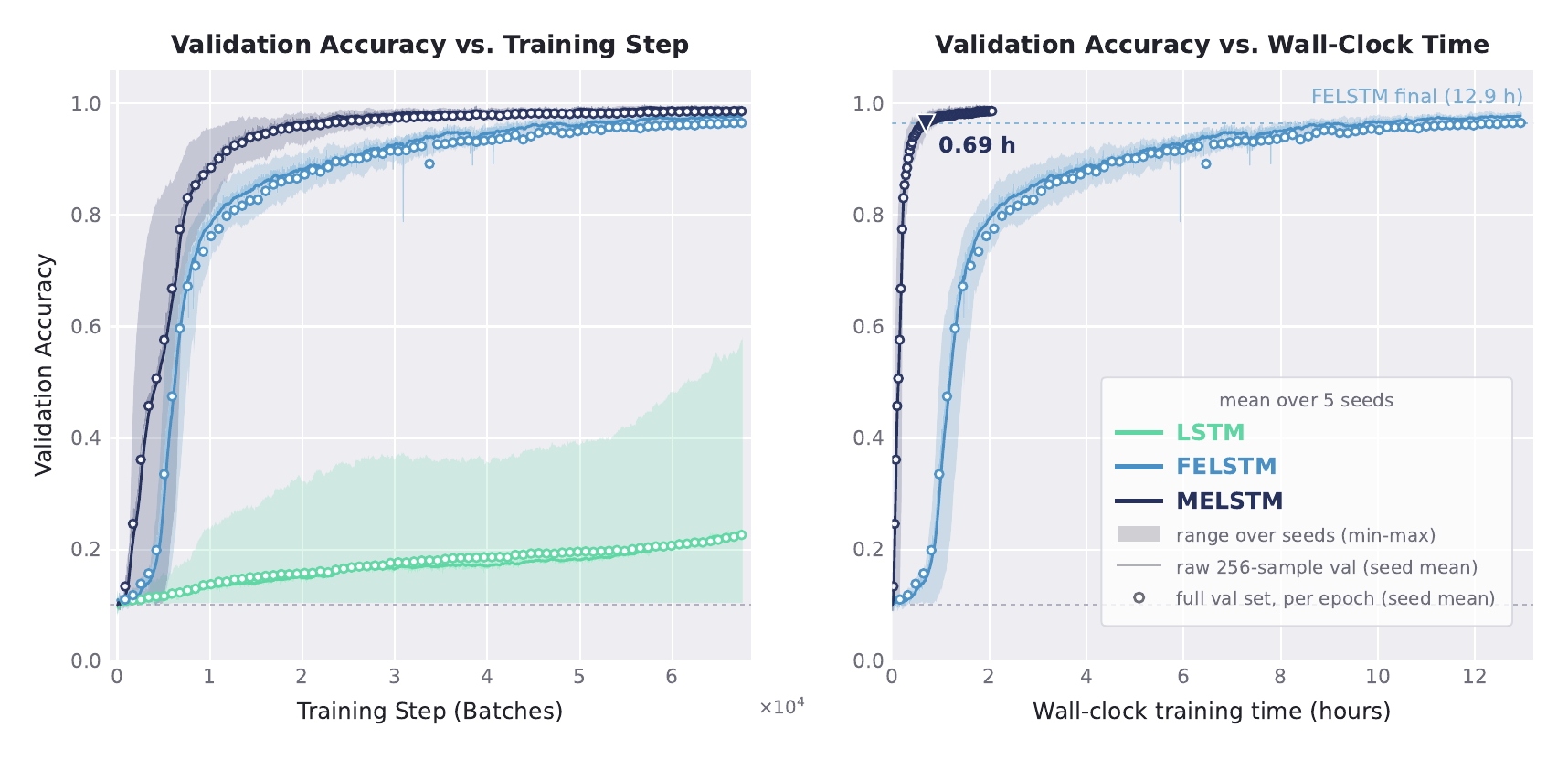}
\end{minipage}\hfill
\begin{minipage}[c]{0.27\linewidth}
  \centering
  \footnotesize
  \setlength{\tabcolsep}{2.5pt}
  \renewcommand{\arraystretch}{1.35}
  \begin{tabular}{lcc}
  \toprule
  Model & \begin{tabular}{@{}c@{}}acc.\ $\uparrow$\\[-1pt]{\tiny val}\\[-3pt]{\tiny test}\end{tabular}
        & \begin{tabular}{@{}c@{}}s /\\epoch\end{tabular} \\
  \midrule
  LSTM   & \begin{tabular}{@{}c@{}}$0.227_{\pm .210}$\\[-2pt]$0.229_{\pm .209}$\end{tabular} & $70_{\pm 4}$ \\
  \addlinespace[2pt]
  FELSTM & \begin{tabular}{@{}c@{}}$0.966_{\pm .009}$\\[-2pt]$0.969_{\pm .008}$\end{tabular} & $582_{\pm 5}$ \\
  \addlinespace[2pt]
  MELSTM & \begin{tabular}{@{}c@{}}$\mathbf{0.987_{\pm .002}}$\\[-2pt]$\mathbf{0.987_{\pm .003}}$\end{tabular} & $93_{\pm 1}$ \\
  \bottomrule
  \end{tabular}
\end{minipage}
\caption{\textbf{Motion-Only MNIST: Training efficiency and test accuracy.}  \textbf{Right:} five-seed mean $\pm$ s.d.\ of the
best validation and the test accuracy and of the median time per epoch. Dotted lines: chance ($10\%$).}
\label{fig:cf_training}
\end{figure}

\begin{figure}[t]
  \centering
  \includegraphics[width=\linewidth]{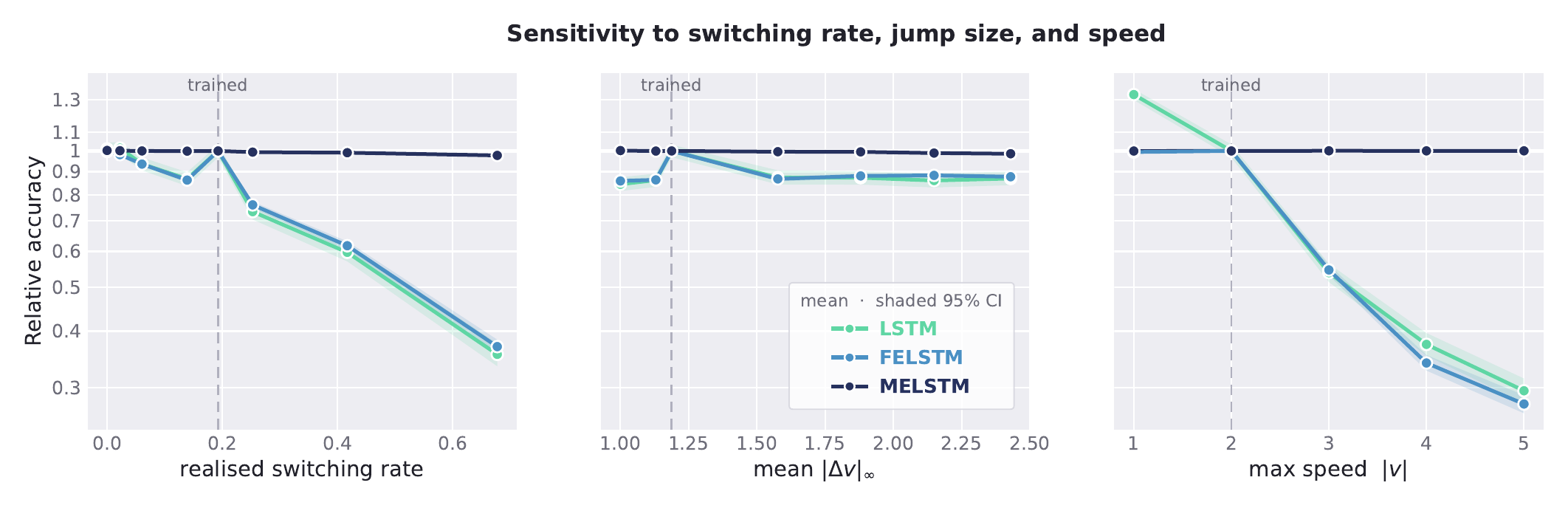}
\caption{\textbf{Motion generalization sensitivity in classification.} Relative to their own training accuracy (far lower for \CL{}),
\CL{} and \FE{} respond alike, except at speed $1$, where \CL{} improves because its $3\times3$ recurrence can move its memory with the digit.}
\label{fig:cf_motion}
\end{figure}

\paragraph{Data and task.}
If every frame is noise, an object can still be defined by the \emph{common
fate} of its pixels, which move together; all information about it then lies in
the correlations between frames~\citep{upadhyay2026timeblindness}.
Motion-Only~MNIST renders a single digit in this way, as a glyph-shaped window
onto its own white-noise texture, which moves with it, over an independent
white-noise background with a motion of its own (Appendix~\ref{app:cf}). No
single frame contains the digit (a single-frame CNN scores $12\%$, against
$10\%$ chance), and the model must name the digit after the context.

\paragraph{Velocity model and training.}
\ME{} reads both slot velocities from the two strongest phase-correlation peaks
of each frame pair $(\CX_{t-1},\CX_t)$, one from the background and one from the digit. The estimator is still parameter-free, and with slot
tracking instead, accuracy is only slightly lower (Appendix~\ref{app:cf}). All
models are trained with cross-entropy for $80$ epochs on five seeds.

\paragraph{Transport reveals the digit.}
A state transported with the digit accumulates its texture, while the
background, moving differently, averages out (Figure~\ref{fig:cf_states}). Both equivariant models exploit this (Figure~\ref{fig:cf_training}),
but \ME{} is slightly more accurate, trains several times faster with $2$ state
copies instead of $25$, and generalizes better (Figure~\ref{fig:cf_motion}).
\CL{} compares consecutive frames within its $3\times3$ recurrence but cannot
carry its memory with a digit moving faster than one pixel per step, so it must
find motion detectors in noise, which it does only on some seeds
(Appendix~\ref{app:cf}).

\paragraph{Motion generalization.}
We evaluate the trained models as the switching rate, jump size and maximum
speed move away from training (Figure~\ref{fig:cf_motion}). As in prediction
(Figure~\ref{fig:motion_generalization}), \FE{} loses accuracy as velocity
changes become frequent, since each change moves its evidence to another
channel, and it does not generalize beyond its velocity grid; \ME{}, which
re-estimates its velocities at every step, generalizes on all three axes.
\CL{}'s $3\times3$ recurrence moves its memory by one pixel per step, so at
speed $1$ it can follow the digit and improves; faster digits leave it only the
comparison of the last two frames, which fails beyond $3$ pixels per frame. Like \FE{}, it also loses accuracy as velocity changes become more frequent,
which suggests that it too accumulates evidence only while the velocity stays
the same.

\subsection{KTH Action Recognition under Camera Motion}
\label{sec:kth}

\begin{figure}[t]
\centering
\includegraphics[width=\linewidth]{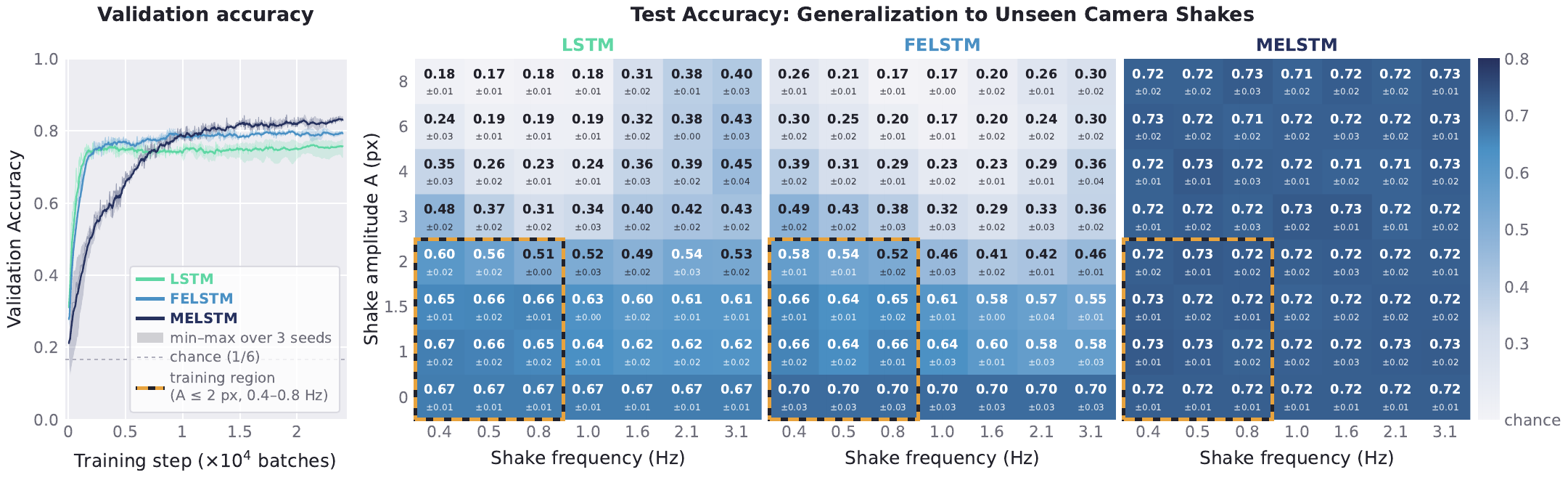}
\caption{\textbf{KTH: training and generalization to camera shake.} Left:
validation accuracy under the training shake. Right: test accuracy under shakes
of amplitude $A$ and frequency $f$ ($A=0$: static camera); mean $\pm$ s.d.\ over
three seeds.}
\label{fig:kth}
\end{figure}

\paragraph{Data and task.}
KTH~\citep{schuldt2004recognizing} contains videos of $25$ people performing six
actions, such as walking, boxing and hand waving, filmed by a mostly static
camera. We simulate a moving camera by translating every frame, which moves the
whole scene but leaves the action unchanged (Appendix~\ref{app:kth}). The models
are trained to recognize the action under a small, slow camera shake, a sinusoid
on each axis with amplitude $A\le2$~px and frequency $f=0.4$--$0.8$~Hz, and tested
under larger and faster shakes and other camera motions. \FE{} enumerates the $9$ velocities with
$\|v\|_\infty\le1$, which contain every step of the training shake. \ME{} has two slots and uses slot tracking, which is equivariant to any camera path;
early in training it also uses frame-pair velocities, with the stochastic
schedule of Appendix~\ref{app:pred_xtrain}. All models are trained for $500$ epochs with cross-entropy loss.

\paragraph{Camera-motion generalization.}
With a static camera ($A=0$) the three models are comparable. Under the training
shake, \ME{} learns more slowly while frame-pair velocities are withdrawn, but ends
with the highest validation accuracy (Figure~\ref{fig:kth}). As the shake grows,
\CL{} and \FE{} lose accuracy, already inside the training region, and approach
chance for large, slow shakes, while \ME{} keeps its static accuracy everywhere,
up to camera steps of $11$ pixels per frame. \FE{} gains little over \CL{},
although its lattice contains every step of the training shake: each of its
copies moves at one fixed velocity, while a shake keeps changing velocity.

\section{Discussion}
\label{sec:discussion}

\paragraph{Related work.}
Most equivariant recurrent and dynamical models are equivariant to static
transformations, applied to every frame alike
\citep{azari2022equivariant,nguyen2023equivariant}. Symmetries that mix space
and time, such as Galilean boosts in fluid dynamics \citep{wang2021incorporating}
and Lorentz boosts in particle physics \citep{bogatskiy2020lorentz}, act on the
whole space-time signal as one group element. FERNNs
\citep{keller2025flowequivariantrecurrentneural} and their extensions to world
models with known self-motion \citep{lillemark2026flow} and to transformers
\citep{khaliliya2026flow} are flow equivariant by construction; \ME{} instead
estimates the motion from the data and is equivariant to motions whose velocity
changes at every step. Recurrent video models such as TrajGRU \citep{shi2017deep}
and MotionRNN \citep{wu2021motionrnn} also move their hidden states along
flow fields that they learn from the data, which improves the prediction of
complex motion. Here we treat motion as a gauge symmetry, under which the
recurrent model is exactly equivariant whenever its transporter is
(Theorem~\ref{thm:motion_equivariance}). Equation~\eqref{eq:velocity_law} states
what this requires of the velocity model, and our experiments measure what the
symmetry buys by comparing models that share an architecture and differ only in
their transport.

\paragraph{Future directions.}
Our symmetry groups act on each frame as a whole or, through slots, on a few
components, each moved by a single group element, and the velocity model in our
experiments is parameter-free. We see three directions that follow from the
construction.

\emph{Trainable velocity models.} Any velocity model that satisfies
\eqref{eq:velocity_law} keeps \ME{} exactly equivariant, including a trainable
one, such as a recurrent network coupled to \ME{}. Phase correlation on
learned convolutional features instead of raw pixels still satisfies it, and
would let the model learn which part of the change between frames is motion and
which is intrinsic (Figure~\ref{fig:motion_vs_flows}). A recurrent velocity
model could also predict the motion after the context.

\emph{Diffeomorphisms.} Atmospheric fields deform as they move, so their motion
is better described by a time-dependent diffeomorphism. For
$G=\mathrm{Diff}(M)$, the transporter advects the states along a velocity
field. Replacing convolutions by diffeomorphism-equivariant
maps, approximated by canonicalization through image registration
\citep{oettinger2026diffeomorphism}, would give a model suited to weather
forecasting.

\emph{Transformers.} Attention from frame $t$ to frame $s$ can read the keys and
values of frame $s$ transported by $T_{t,s}$. By
\eqref{transporter_equivariance_relation} these transform with $\psi_t$, like the
query, which would extend flow equivariant transformers
\citep{khaliliya2026flow} from flows to motions.

\ifdeanonymized
\subsubsection*{Acknowledgments}
V.H. and M.P. received support from National Science Foundation Award \#2418909 LEAPS-MPS: Topological Machine Learning and Cryo-EM.
T.A.K. acknowledges that this work has been made possible in part by a gift from the Chan Zuckerberg Initiative Foundation to establish the Kempner Institute for the Study of Natural and Artificial Intelligence at Harvard University. M.P. recieved support from CAPITAL Services.
\fi

\bibliographystyle{tmlr}
\bibliography{references}

\newpage

\appendix

\addtocontents{toc}{\protect\setcounter{tocdepth}{3}}
\renewcommand{\contentsname}{Appendix Contents}
\tableofcontents

\section{Proofs}

\subsection{The structure of subgroups of \texorpdfstring{$\mathcal{G}$}{G}}\label{structure_of_subgroups_of_G}
\begin{lemma}\label{lemma-flows-subgroup}
The motions form a subgroup of $\mathcal{G}$, while the flows form a subgroup of $\mathcal{G}$ if and only if $G$ is an abelian group.
\end{lemma}

\begin{proof}
We assume that $\psi^{(1)}$ and $\psi^{(2)}$ are two motions generated by the velocities $v^{(1)}$ and $v^{(2)}$, respectively. We show that $\psi_t = \psi^{(1)}_t \psi^{(2)}_t$ is again a motion. Differentiating with respect to time and using the product rule gives $\frac{d}{dt}\psi_t = \dot{\psi}^{(1)}_t \psi^{(2)}_t + \psi^{(1)}_t \dot{\psi}^{(2)}_t$. Substituting the evolution equations for $\psi^{(1)}$ and $\psi^{(2)}$ yields
\begin{align}
\frac{d}{dt}(\psi^{(1)}_t \psi^{(2)}_t) = \dot{\psi}_t &= v^{(1)}_t \psi^{(1)}_t \psi^{(2)}_t+\psi^{(1)}_t v^{(2)}_t \psi^{(2)}_t\\
&= v^{(1)}_t \psi^{(1)}_t \psi^{(2)}_t+\psi^{(1)}_t v^{(2)}_t {\psi^{(1)}_t}^{-1} \psi^{(1)}_t\psi^{(2)}_t
\end{align}
Using the definition of the adjoint action $\operatorname{Ad}_g(v) = gvg^{-1}$ for $g \in G$ and $v \in \mathfrak g$, we conclude that
$\psi_t$ is again a motion generated by the velocity
\begin{align}
    v_t = v^{(1)}_t + \operatorname{Ad}_{\psi^{(1)}_t} (v^{(2)}_t)
\end{align}
which is simply $v_t = v^{(1)}_t + v^{(2)}_t$ in the case of an abelian group where the adjoint action is trivial.
Similarly, we can show that the inverse of a motion is again a motion, and therefore the motions form a subgroup of $\mathcal{G}$. This is proved as follows. Starting with the identity $\psi_t^{-1} \psi_t = e$ and differentiating gives $\dot{\psi}_t^{-1} \psi_t + \psi_t^{-1} \dot{\psi}_t = 0$. Substituting $\dot{\psi}_t = v_t \psi_t$ yields $\dot{\psi}_t^{-1} \psi_t + \psi_t^{-1} v_t \psi_t = 0$. Multiplying on the right by $\psi_t^{-1}$ gives $\dot{\psi}_t^{-1} + \psi_t^{-1} v_t = 0$, so $\dot{\psi}_t^{-1} = -\psi_t^{-1} v_t$. Rewriting in right-invariant form gives $\dot{\psi}_t^{-1} = -\operatorname{Ad}_{\psi_t^{-1}}(v_t) \psi_t^{-1}$, showing that $\psi_t^{-1}$ is a motion generated by the velocity $v^{\mathrm{inv}}_t = -\operatorname{Ad}_{\psi_t^{-1}}(v_t)$.

For the second part of the lemma we proceed as follows. For the forward direction, if $G$ is abelian then the composition of two flows $\phi^{(1)}_t = \exp(t \nu_1)$ and $\phi^{(2)}_t = \exp(t \nu_2)$ is given by $\phi_t = \phi^{(1)}_t \phi^{(2)}_t = \exp(t (\nu_1 + \nu_2))$, which is again a flow. The inverse of a flow $\phi_t = \exp(t \nu)$ is given by $\phi_t^{-1} = \exp(-t \nu)$, which is also a flow. Thus, if $G$ is abelian, the flows form a subgroup of $\mathcal{G}$. For the converse direction, if the flows form a subgroup of $\mathcal{G}$, then for any $\nu_1, \nu_2 \in \mathfrak g$, the composition of the corresponding flows $\phi^{(1)}_t = \exp(t \nu_1)$ and $\phi^{(2)}_t = \exp(t \nu_2)$ must be a flow. This implies that $\exp(t \nu_1) \exp(t \nu_2) = \exp(t \nu)$ for some $\nu \in \mathfrak g$.
Differentiating at $t=0$ gives $\nu = \nu_1 + \nu_2$. Using the Baker-Campbell-Hausdorff formula,
$\exp(t\nu_1)\exp(t\nu_2) = \exp\Big(t(\nu_1+\nu_2) + \tfrac{t^2}{2}[\nu_1,\nu_2] + \mathcal{O}(t^3)\Big)$ shows then that equality for all $t$ is possible only if $[\nu_1, \nu_2] = 0$ for all $\nu_1, \nu_2 \in \mathfrak g$. Therefore, the Lie algebra $\mathfrak g$ is abelian, and hence the Lie group $G$ is abelian.
\end{proof}

\paragraph{Group structure of ${\cal G}$.}
Consider the subgroup ${\cal G}_0 = \{\gamma \in {\cal G} : \gamma_0 = e\}$ consisting of time-dependent transformations initialized at the identity. This is a normal subgroup of ${\cal G}$ and satisfies ${\cal G}_0 \cap {\cal G}_{glob} = \{e\}$. Moreover, every element $\gamma \in {\cal G}$ admits a unique decomposition into a component in ${\cal G}_0$ and a global component in ${\cal G}_{glob}$, since $\gamma_t \gamma_0^{-1} \in {\cal G}_0$ and $\gamma_0 \in {\cal G}_{glob}$. Hence, every $\gamma \in {\cal G}$ can be written uniquely as
$\gamma = (\gamma \gamma_0^{-1})\gamma_0$, which gives the internal product decomposition
$\cG = \cG_0 \cG_{glob}$.
Therefore, by \citet[Theorem 12, Section 5.5]{dummit2004abstract}, $\cG = \cG_0 \rtimes \cG_{glob}$.
The semidirect product structure is determined by the conjugation action of ${\cal G}_{glob}$ on ${\cal G}_0$, given by ${(g \cdot \gamma)}_t = g \gamma_t g^{-1}$ for $g \in {\cal G}_{glob}$ and $\gamma \in {\cal G}_0$. Since motions and flows satisfy $\psi_0 = e$ and $\phi_0 = e$, respectively, they are naturally subsets of ${\cal G}_0$. Consequently,
${\cal G}_{flow} \subset {\cal G}_{mot} \le {\cal G}_0 \vartriangleleft {\cal G}$,
and ${\cal G}_{glob} \le {\cal G}$.

\subsection{Global G-equivariance of \CL{} model}\label{G-equivariance_of_ConvLSTM_model}
\begin{theorem}[\CL{} is globally $G$-equivariant]
Consider the \CL{} model defined by equations~\eqref{ConvLSTM_equations_1}--\eqref{ConvLSTM_equations_6}, with invariant initial state variables satisfying $g \triangleright \CH_0 = \CH_0$ and $g \triangleright \CC_0 = \CC_0$ for all $g \in {\cal G}_{glob}$. Then the model is globally $G$-equivariant, i.e., for any global transformation $g \in {\cal G}_{glob}$ and any trajectory $\CX \in {\cal F}$,
\begin{align}
        \CH_{t}[g \triangleright \CX_{\le t}] = g \triangleright \CH_{t}[\CX_{\le t}] \qquad \forall t.
\end{align}
\end{theorem}

\begin{proof}
We want to prove
\begin{align}
    {\cal H}_{t}[g \triangleright \CX_{\le t}] = g\triangleright {\cal H}_{t}[\CX_{\le t}] \qquad \forall t, \qquad \forall g \in \cG_{glob}
\end{align}
The proof is by induction on $t$. For the base case, we consider the initial states ${\cal C}_0$ and ${\cal H}_0$ to be invariant under the group action, so $g \triangleright \CC_0(x) = \CC_0(x)$ and $g \triangleright \CH_0(x) = \CH_0(x)$. Moreover, the empty sequence $\CX_{<0}$ is trivially invariant under the group action, so we have the equivariance at time $t=0$ as follows $g \triangleright \CH_0[\CX_{<0}] = \CH_0[\CX_{<0}] = \CH_0[g \triangleright \CX_{<0}]$ (similarly for ${\cal C}_0$). Now, assuming that the equivariance holds at time $t$ we want to show that it also holds at time $t+1$. First note that for the gate variables we have
\begin{align}
    z_t[g \triangleright \CX_{\le t}](x)&= a_z\Big[ [W_{z} \star {\cal H}_t[g \triangleright \CX_{<t}]](x) + [U_{z} \star (g\triangleright {\cal X}_t)](x)\Big] \\
    &= a_z\Big[ [W_{z} \star (g \cdot {\cal H}_t[ \CX_{<t}])](x) + [U_{z} \star (g\cdot {\cal X}_t)](x) \Big]\\
    &= a_z \Big[g \cdot[W_{z} \star {\cal H}_t[ \CX_{<t}]](x) + g \cdot [U_{z} \star  {\cal X}_t](x)  \Big]\\
    &= g \triangleright z_t[\CX_{\le t}](x)
\end{align}
for all $z \in \{i,f,o,c\}$. Here in the second line, we used the induction hypothesis for the hidden state and in third line the equivariance of the steerable convolution under regular action of the group.
For the cell state we have
\begin{align}
    \CC_{t+1}[g \triangleright \CX_{\le t}](x) &= {\cal C}_t[g \triangleright \CX_{< t}](x)  \odot  f_t[g \triangleright \CX_{\le t}](x)  + i_t[g \triangleright \CX_{\le t}](x)  \odot c_t[g \triangleright \CX_{\le t}](x) \\
    &= [g \cdot {\cal C}_t[ \CX_{< t}]](x) \odot  [g \cdot f_t[ \CX_{\le t}]  ](x) + [g \cdot i_t[ \CX_{\le t}]](x)  \odot [g \cdot c_t[ \CX_{\le t}] ](x)\\
    & = g \cdot [ {\cal C}_t[ \CX_{< t}]  \odot f_t[ \CX_{\le t}] ](x) + g \cdot [ i_t[ \CX_{\le t}]  \odot  c_t[ \CX_{\le t}] ](x)\\
    &= g \triangleright \CC_{t+1}[ \CX_{\le t}](x)
\end{align}
where in the second line we used the induction hypothesis for the cell state and the gate variables and in the third line we used the fact that the group action distributes over the Hadamard product.
Finally for the hidden state we have
\begin{align}
    \CH_{t+1}[g \triangleright \CX_{\le t}](x) &= o_t[g \triangleright \CX_{\le t}](x)  \odot \tanh ({\cal C}_{t+1}[g \triangleright   \CX_{\le t}](x) ) \\
    &= g \cdot o_t [ \CX_{\le t}](x)  \odot \tanh (g \cdot {\cal C}_{t+1}[  \CX_{\le t}](x) )\\
    &= g \triangleright \CH_{t+1}[ \CX_{\le t}](x)
\end{align}
where in the second line we used the induction hypothesis for the gate variable $o$ and the cell state and in the third line we used the fact that the group action distributes over the Hadamard product and also commutes with the pointwise nonlinearity $\tanh$.
\end{proof}

\subsection{\CL{} is not flow equivariant}\label{flow_equivariance_implies_constant_kernels_for_DenseLSTM}
\begin{proposition}\label{prop:noneq}
Let $\cal N$ be a \CL{} defined by equations~\eqref{ConvLSTM_equations_1}--\eqref{ConvLSTM_equations_6}, with invariant initial states
$\CH_0=\CC_0=0$ whose kernels are supported in a ball of radius $R$, on $\R^2$ or on a torus of width $L>8R$. If $\cal N$ is flow equivariant, then
\begin{equation}\label{eq:C_mem=0}
\CC_1\odot\sigma\bigl(W_f\star\CH_1\bigr)
+\sigma\bigl(W_i\star\CH_1\bigr)\odot\tanh\bigl(W_c\star\CH_1\bigr) = \;0,
\qquad
    \begin{aligned}
    \CC_1&=\sigma(U_i)\odot\tanh(U_c),\\
    \CH_1&=\sigma(U_o)\odot\tanh(\CC_1).
    \end{aligned}
\end{equation}
Condition~\eqref{eq:C_mem=0} fails for all kernels outside a Lebesgue measure zero set (e.g.\ it
fails whenever $W_i=W_f=W_c=0$ and $U_c\neq0$); hence $\cal N$ is not flow equivariant for
generic kernels.
\end{proposition}

\begin{proof}
As a probe for flow equivariance, consider the sequence of static dots $\CX_t=\delta_0$ for all $t$ where a flow of velocity $\nu$ maps it to the moving dot $(\phi\triangleright\CX)_t=\delta_{t\nu}$. Since $(U\star\delta_a)(x)=U(x-a)$ and $\CC_0=\CH_0=0$, the first time step gives $\CC_1=\sigma(U_i)\tanh(U_c)$ and $\CH_1=\sigma(U_o)\tanh(\CC_1)$, supported in the ball $B_R$ (they vanish where $U_c$ does as $\sigma(x) > 0$ and $\tanh(x) = 0$ iff $x = 0$), so the memory terms $P_z:=W_z\star\CH_1$ are supported
in $B_{2R}$. Choose $4R<|\nu|<L-4R$ and evaluate the equivariance condition at time $t=1$,
\[
\CH_2\bigl[(\delta_0,\delta_\nu)\bigr](x)=\CH_2\bigl[(\delta_0,\delta_0)\bigr](x-\nu),
\]
at points $|x|\le2R$. On the left, all time-$1$ input terms $U_z(x-\nu)$ vanish, since
$|x-\nu|\ge|\nu|-|x|>2R\ge R$; the gates see only their memory, leaving
$\sigma(P_o)\,\tanh\bigl(\CC_1\,\sigma(P_f)+\sigma(P_i)\tanh(P_c)\bigr)(x)$. On the right,
every argument of the static response vanishes at $x-\nu$ (distance $>2R$ from the dot),
giving $\sigma(0)\tanh(0)=0$. Since $\sigma>0$ and $\tanh(s)=0$ only at $s=0$, so the memory content $\CC_2^{mem} =\CC_1\,\sigma(P_f)+\sigma(P_i)\tanh(P_c) $ must vanish on  $|x|\le2R$ where it is also zero trivially outside of $B_{2R}$. This gives~\eqref{eq:C_mem=0} as a necessary condition for flow equivariance.

In practice, the convolution kernels are finite $k\times k$ arrays, so all parameters can be collected into a vector in a finite-dimensional space $\theta\in\R^D$, where $D=8k^2$. Since $\CC_2^{mem}$ is a real-analytic function of the kernel
parameters and it is not identically zero (as it is not zero whenever $U_c(x)\neq0$), it follows that its zero set has Lebesgue measure zero in
$\R^D$. Since every flow-equivariant \CL{} must satisfy
\eqref{eq:C_mem=0}, the set of flow-equivariant kernels is contained in
this measure-zero set.
\end{proof}

\begin{remark}[Space-time convolutions]\label{rem:spacetime}
Viewing a sequence as a field $\CX(x,t)$ on space-time, a linear map equivariant
to space-time translations is a space-time convolution,
$\Phi[\CX](x,t)=\sum_{s}\int K(x-y,\,t-s)\,\CX(y,s)\,dy$. A flow of translations
with velocity $\nu$ acts on space-time as the Galilean boost
$(x,t)\mapsto(x+\nu t,t)$, and $\Phi$ commutes with all such boosts if and only if
\begin{align}\label{eq:boost_constraint}
K(x+\nu\tau,\,\tau)=K(x,\tau)\qquad \forall\, \nu,x\in\R^2,\ \forall\,\tau .
\end{align}
For $\tau\neq0$ the orbit $\{x+\nu\tau:\nu\in\R^2\}$ is the whole plane, so
$K(\cdot,\tau)$ is constant: such a map couples different times only through
spatial averages, and with a local kernel it does not couple them at all. For
fields that transform under the boosts by a finite-dimensional representation
$\rho$, \eqref{eq:boost_constraint} becomes
$K(x+\nu\tau,\tau)=\rho_{out}(\nu)K(x,\tau)\rho_{in}(\nu)^{-1}$, which still
determines $K(\cdot,\tau)$ by its value at $x=0$. Lifting the fields to the
boosts themselves, an infinite-dimensional representation, avoids the
constraint; this is the route of FERNN and \FE{}. The orbits of Lorentz boosts,
in contrast, are the hyperboloids $\tau^2-|x|^2=\mathrm{const}$ (with $c=1$),
which meet each time slice in a circle, so a Lorentz invariant kernel
$K(x,\tau)=k(\tau^2-|x|^2)$ can still vary in space.
\end{remark}

\subsection{Transport and flow convolution}\label{app:transport_convolution}
In this section we introduce the transport convolution and the transport-lifting convolution: the $\cG$-equivariant linear maps between transport fields on $M \times G$, and from regular fields on $M$ to transport fields on $M \times G$, respectively. As special cases we then obtain the flow-lifting convolution $\hat{\star}$ and the flow convolution $\star_F$, the two operations used in the \FE{} architecture.
Throughout, we assume that $G$ is unimodular, so that its Haar measure is both left and
right invariant. For $G = \R^2 \rtimes H$, unimodularity is equivalent to $|\det h| = 1$ for
all $h \in H$; in that case $G$ also preserves the Lebesgue measure on $M = \R^2$, so all
Jacobian factors $|\det J_k|$ appearing below are equal to one.

\paragraph{Transport-lifting convolution.} We begin with a general linear map that takes as input a regular field on $M$ and produces a transport field on $M \times G$:
\begin{align}
{\cal I}_U[\CX_t](x,g) = \int_M U(x,y;g) \CX_t(y) dy
\end{align}
where $U : M \times M \times G \to \R^{c_{out}\times c_{in}}$ is the kernel. Requiring this map to be equivariant means that ${\cal I}_U[\gamma\triangleright \CX_t] = \gamma\triangleright {\cal I}_U[\CX_t]$ for all $\gamma \in {\cal G}$. Expanding both sides gives
\begin{align}
    \int_M U(x,y;g) \CX_t(\gamma_t^{-1} \cdot y) dy = \int_M U(\gamma^{-1}_t \cdot x, y; \gamma^{-1}_t g \gamma_{t-1}) \CX_t(y) dy
\end{align}
for all $\gamma \in {\cal G}$, all $\CX \in {\cal F}$ and all $t \ge 0$.
Using the change of variables $y' = \gamma_t^{-1} \cdot y$, the left-hand side becomes
$\int_M U(x,\gamma_t \cdot y';g) \CX_t(y') |\det J_{\gamma_t}| dy' $, where $J_{\gamma_t}$ is the Jacobian $\frac{\partial y}{\partial y'}$ of the transformation $\gamma_t\in G$. Since the above equality must hold for all inputs $\CX\in{\cal F}$, we obtain the kernel constraint
\begin{align}
    U(\gamma_t \cdot x,\gamma_t \cdot y; \gamma_t g \gamma_{t-1}^{-1}) = \frac{1}{|\det J_{\gamma_t}|} U(x,y;g)
\end{align}
which must hold for all $\gamma \in {\cal G}$.
Taking $t=0$, where $\gamma_0 = e$ for $\gamma \in \cG_0$, the spatial factors drop out and the constraint reduces to
$U(x,y;\, g\,\gamma_{-1}^{-1}) = U(x,y;g)$ for all $\gamma_{-1} \in G$. Since the right action of
$G$ on itself is transitive, it follows that $U$ must be independent of its group argument, i.e.\
$U(x,y;g) = U(x,y)$. The remaining constraint is precisely the usual equivariance condition on
the spatial variables, implying that $U$ is a steerable convolution kernel. In particular, global transformations $\gamma \in \cG_{glob}$ impose no
further condition, since $U$ no longer carries a group argument on which conjugation could act. Conversely, any kernel satisfying these conditions defines a $\cG$-equivariant lifting operator. Hence the above characterization of $U$ is both necessary and sufficient for $\cG$-equivariance, and the lifting map reduces to the standard steerable convolution
\begin{align}
    [U \, \hat{\star}\, \CX_t](x,g) = \int_M U(x-y) \CX_t( y) dy = [U \star \CX_t](x)
\end{align}
where we copy the result into the extra channels indexed by $G$. Note that this map is equivariant under the full group $\cG$, including the flow, motion and global transformations.

\paragraph{Transport convolution.} Following the same logic, we define the transport convolution as a $\cG$-equivariant linear map between transport fields on $M \times G$. A general such map takes the form
\begin{align}
    {\cal I}_W[\CH_t](x,g) = \int_M \int_G W(x,y;g,{\tilde g}) \CH_t(y,{\tilde g})  \, d{\tilde g}\, dy
\end{align}
for a kernel $W: M \times M \times G \times G \to \R^{c_{out} \times c_{in}}$ and the
Haar measure $d{\tilde g}$ on $G$. The equivariance condition ${\cal I}_W[\gamma \triangleright \CH_t] = \gamma\triangleright
{\cal I}_W[\CH_t]$ results in the constraint
\begin{align}
    W(\gamma_{t-1} \cdot x, \gamma_{t-1} \cdot y;\, \gamma_t g \gamma_{t-1}^{-1},\,
    \gamma_t {\tilde g} \gamma_{t-1}^{-1}) = \frac{1}{|\det J_{\gamma_{t-1}}|} W(x,y;g,{\tilde g})
\end{align}
for all $\gamma \in {\cal G}$, all $x,y \in M$ and all $g,{\tilde g} \in G$.
Again, a necessary condition is obtained by taking $t=1$, which gives
$W(x, y; kg, k{\tilde g}) = W(x,y;g,{\tilde g})$ for all $k,g,{\tilde g} \in G$. Setting $k = {\tilde g}^{-1}$ shows
that the kernel depends on the group arguments only through ${\tilde g}^{-1}g$, i.e.\
$W(x,y;g,{\tilde g}) = W(x,y;\,{\tilde g}^{-1}g)$, so the integration over $G$ takes the form of a
group convolution. For $t=2$ the constraint becomes
\begin{align}\label{eq:jointW}
    W(k \cdot x,k \cdot y;\,k q k^{-1}) =  \frac{1}{|\det J_{k}|} W(x,y;q)
\end{align}
for all $k,q \in G$, with $q = {\tilde g}^{-1}g$ and $k = \gamma_1$; the constraints at $t \geq 3$ follow by iterating this argument and impose no additional structure. Since the same $k$ acts on the spatial and on the group argument, \eqref{eq:jointW} is a single joint condition and does not in general factor into a spatial equivariance condition and a separate conjugation invariance.
Equivalently, $W$ is an equivariant function on $M \times M \times G$ under the diagonal action $(x,y,q)\mapsto(k\cdot x,\,k\cdot y,\,kqk^{-1})$, and is therefore freely specifiable on a set of orbit representatives.
Rather than solving this condition in general, we restrict to the case used in the \FE{}, in which the bank is an abelian subgroup $V \le G$. Transport fields then live on $M \times V$, and the relevant transformations are those taking values in $V$, namely the motions ${\cal V}_{mot}$ and in particular the flows ${\cal V}_{flow}$. Conjugation by an element of $V$ acts trivially on $V$, so the velocity arguments are only translated, and~\eqref{eq:jointW} restricted to $k \in V$ reduces to invariance of $W$ under a common spatial translation. This gives $W(x,y;q) = W(x-y;q)$ and hence~\eqref{flowConvolution}: a convolution in both arguments, where we write $\mathsf{v}-\mathsf{u}$ for $\mathsf{u}^{-1}\mathsf{v}$ since $V$ is abelian. In particular, $\star_F$ is equivariant under all of ${\cal V}_{mot}$, and not only under ${\cal V}_{flow}$.
Note that if we keep the same domain $M \times V$ but also want the operation to be equivariant under global transformations, that is, under ${\cal V}_{mot} \rtimes \cG_{glob}$, then two further conditions are needed. First, $\cG_{glob}$ must act on ${\cal V}_{mot}$ by conjugation, so the action $\mathsf{v}\mapsto k\mathsf{v}k^{-1}$ of $k \in G$ on the bank must land in $V$; that is, $V$ must be normal in $G$. Second, \eqref{eq:jointW} must now hold for all $k \in G$, and not only for $k \in V$. For $G = \R^2 \rtimes H$ with $V = \R^2$ both conditions are satisfied, and the conjugation can be computed explicitly: writing $k = (\tau,h)$ and $\mathsf{v} = (b,I)$,
\begin{align}
    k\,\mathsf{v}\,k^{-1} = (\tau,h)(b,I)(-h^{-1}\tau,\,h^{-1}) = (\tau + hb,\; h)(-h^{-1}\tau,\, h^{-1}) = (h b,\; I),
\end{align}
so $V$ is indeed normal and conjugation acts on the bank through the point group alone, $\mathsf{v}\mapsto h\,\mathsf{v}$, independently of $\tau$. This is why the analysis above is unaffected: translations act only on the spatial argument. The remaining $h$-dependence in~\eqref{eq:jointW} then yields the joint steerability constraint
\begin{align}
    W(h \cdot z;\; h\,q) = W(z;\,q), \qquad \forall\, h \in H.
\end{align}

\subsection{Flow equivariance of \FE{}}\label{channel_based_FEConvLSTM_equivariance}
In this section, we prove Theorem~\ref{thm:flow_equivariance} which says that the channel based \FE{} is flow equivariant.

\begin{proof}
The proof proceeds by induction on $t$.
The base case $t=0$ is immediate since, by assumption, the initial state variables $\CH_0$ and $\CC_0$ are invariant under the action of flows, i.e.,
$\phi \triangleright \CH_0 = \CH_0$ and $\phi \triangleright \CC_0 = \CC_0$ for all $\phi \in {\cal V}_{flow}$ where $V \le G$ is an abelian subgroup of $G$. Moreover, the empty sequence $\CX_{<0}$ is trivially invariant under the action of flows. Therefore,
\begin{align}
    \phi \triangleright \CH_0[\CX_{<0}] = \CH_0[\CX_{<0}] = \CH_0[\phi \triangleright \CX_{<0}]
\end{align}
and similarly for $\CC_0$, establishing the base case of the induction.
For the inductive step, assume that the state variables $\CH_t$ and $\CC_t$ are flow equivariant at time $t$, namely,
$\phi \triangleright \CH_t[\CX_{<t}] = \CH_t[\phi \triangleright \CX_{<t}]$ and
$\phi \triangleright \CC_t[\CX_{<t}] = \CC_t[\phi \triangleright \CX_{<t}]$
for all $\phi \in {\cal V}_{flow}$. We now show that the state variables at time $t+1$ are also flow equivariant.

We first show that the gate variables are flow equivariant:
\begin{align}
    z_t[\phi \triangleright \CX_{\le t}](x,\mathsf{v})
    &= a_z\Big[
        \mathsf{v}\cdot \left[W_{z} \star_F {\cal H}_t[\phi \triangleright \CX_{<t}]\right](x,\mathsf{v})
        + [U_{z} \, \hat{\star} \, {(\phi \triangleright {\cal X})}_t](x,\mathsf{v})
    \Big] \\
    &= a_z\Big[
        \mathsf{v} \phi_{t-1} \cdot
        \left[W_{z} \star_F {\cal H}_t[\CX_{<t}]\right]
        \!\big(x,
            \phi_t^{-1} \mathsf{v}\, \phi_{t-1}
        \big)
        + \phi_t \cdot
        [U_{z} \,\hat\star\, {\cal X}_t]
        \!\big(x,
            \phi_t^{-1} \mathsf{v}\, \phi_{t-1}
        \big)
    \Big] \\
    &=\phi_t \cdot a_z\Big[
            \phi_t^{-1} \mathsf{v}\, \phi_{t-1}
        \cdot
         \left[W_{z} \star_F {\cal H}_t[\CX_{<t}]\right]
        \!\big(x,
            \phi_t^{-1} \mathsf{v}\, \phi_{t-1}
        \big)
        +
        [U_{z} \,\hat\star\, {\cal X}_t]
        \!\big(x,
            \phi_t^{-1} \mathsf{v}\, \phi_{t-1}
        \big)
    \Big] \\
    &= \phi_t \cdot
    z_t[\CX_{\le t}]
    \!\big(x,
        \phi_t^{-1} \mathsf{v} \phi_{t-1}
    \big) \\
    &= \phi \triangleright z_t[\CX_{\le t}](x,\mathsf{v})
\end{align}
for all $z \in \{i,f,o,c\}$. Here, in the first line, we used the definition of the gate variables. In the second line, we applied the inductive hypothesis for the hidden state variable together with the flow equivariance of the flow convolution and flow-lifting convolution operators. In the third line, we inserted the identity element in the form $\phi_t \phi_t^{-1} = e$ into the first term and factored out $\phi_t$, using the fact that the nonlinearity $a_z$ acts pointwise and is equivariant under the action of group elements. In the fourth line,  we used the definition of the gate variables once more, and in the final line we used the definition of the action of flows on transport fields.

Next we show that the cell state $\CC_{t+1}$ equation is flow equivariant as well. Starting with the left hand side, we have
\begin{align}
    \CC_{t+1}[\phi \triangleright \CX_{\le t}](x,g)
    &= g \cdot {\cal C}_t[\phi \triangleright \CX_{< t}](x,g)
    \odot f_t[\phi \triangleright \CX_{\le t}](x,g)
    + i_t[\phi \triangleright \CX_{\le t}](x,g)
    \odot c_t[\phi \triangleright \CX_{\le t}](x,g) \\
    &= g \phi_{t-1} \cdot {\cal C}_t[\CX_{< t}] \big(x,\phi_t^{-1} g\, \phi_{t-1}\big)
    \odot
    \phi_t \cdot
    f_t[\CX_{\le t}]\big(x,\phi_t^{-1} g\, \phi_{t-1}\big)
    \nonumber \\
    & \hspace{3cm}
    + \phi_t \cdot i_t[\CX_{\le t}]\big(x,\phi_t^{-1} g\, \phi_{t-1}\big)
    \odot
    \phi_t \cdot c_t[\CX_{\le t}]
    \big(x, \phi_t^{-1} g\, \phi_{t-1}\big) \\
    &= \phi_t \cdot
    \Big[\phi_t^{-1} g\, \phi_{t-1} \cdot {\cal C}_t[\CX_{< t}]
        \big(x, \phi_t^{-1} g\, \phi_{t-1}\big)
        \odot
        f_t[\CX_{\le t}]
        \big(x, \phi_t^{-1} g \phi_{t-1}\big)
        \nonumber \\
        &\hspace{3cm}
        + i_t[\CX_{\le t}]
        \big(x, \phi_t^{-1} g\, \phi_{t-1}\big)
        \odot
        c_t[\CX_{\le t}]
        \big(x, \phi_t^{-1} g\, \phi_{t-1}\big)
    \Big]\\
    &= \phi_t \cdot
    \CC_{t+1}[\CX_{\le t}]
    \!\big(x,
        \phi_t^{-1} g\, \phi_{t-1}
    \big)\\
    &= \phi_t \cdot
    \CC_{t+1}[\CX_{\le t}]
    \!\big(x,
        \phi_{t+1}^{-1} g\, \phi_{t}
    \big)\\
     &= \phi \triangleright \CC_{t+1}[\CX_{\le t}](x, g)
\end{align}
where in the second line we used the inductive hypothesis for the cell state at time $t$ variable together with the flow equivariance of the gate variables which we just proved.
In the third line, we inserted the identity element in the form $\phi_t \phi_t^{-1} = e$ into the first term and factored out $\phi_t$, using the fact that regular action distributes over the Hadamard product. In the fourth line, we used the definition of the cell state variable at time $t+1$. In the fifth line we used that $\phi$ is a flow in the abelian group $V$, so $\phi_t^{-1} g\, \phi_{t-1} = g\,\phi_{1}^{-1} = \phi_{t+1}^{-1} g\, \phi_{t}$; this is the only step that uses the flow property. In the final line we used the definition of the action of flows on transport fields.

Finally we show that the hidden state $\CH$ is flow equivariant. Starting with the left hand side, we have
\begin{align}
    \CH_{t+1}[\phi \triangleright \CX_{\le t}](x, g)
    &= o_t[\phi \triangleright \CX_{\le t}](x, g)
    \odot
    \tanh \Big(\CC_{t+1}[\phi \triangleright \CX_{\le t}](x, g)\Big) \\
    &= \phi_t \cdot
    o_t[\CX_{\le t}]
    \!\left(x,
        \phi_t^{-1} g \phi_{t-1}
    \right)
    \odot
    \tanh \Big(
        \phi_t \cdot
        \CC_{t+1}[\CX_{\le t}]
        \big(x,
            \phi_t^{-1} g \phi_{t-1}
        \big)
    \Big)\\
    &= \phi_t \cdot
    \CH_{t+1}[\CX_{\le t}]
    \!\left(x,
        \phi_t^{-1} g \phi_{t-1}
    \right)\\
    &= \phi_t \cdot
    \CH_{t+1}[\CX_{\le t}]
    \!\left(x,
        \phi_{t+1}^{-1} g \phi_{t}
    \right)\\
    &= \phi \triangleright \CH_{t+1}[\CX_{\le t}](x,g)
\end{align}
where in the second line we used the flow equivariance of the output gate variable which we proved earlier together with the flow equivariance of the cell state variable which we also proved. In the third line, we used the fact that $\tanh$ is a pointwise nonlinearity and is equivariant under the action of group elements. We also factored out $\phi_t$ using the fact that the regular action distributes over the Hadamard product, and used the definition of the hidden state variable at time $t+1$. In the fourth line we used the flow property, exactly as for the cell state, and in the final line we used the definition of the action of flows on transport fields. This completes the inductive step and hence the proof of flow equivariance for the channel based \FE{} model.
\end{proof}

\paragraph{Global and motion equivariance of the \FE{}.}
Since the flow $\phi$ takes values in the abelian subgroup $V$, the transport conjugation
could equally well have been written in the reduced form
$\phi_t^{-1}\mathsf{v}\,\phi_{t-1} = \mathsf{v}\,\phi_1^{-1}$. We deliberately kept the
unreduced form, because it isolates the single place where the proof uses that $\phi$ is a
flow, and therefore shows exactly why the model fails to be motion equivariant.

The induction produces a field whose velocity argument is $\psi_t^{-1}\mathsf{v}\,\psi_{t-1}$,
whereas the transport-field law at the next time step requires
$\psi_{t+1}^{-1}\mathsf{v}\,\psi_{t}$. Closing the induction therefore requires
\begin{align}
    \psi_t^{-1}\,\mathsf{v}\,\psi_{t-1} = \psi_{t+1}^{-1}\,\mathsf{v}\,\psi_{t}
    \qquad \forall\, \mathsf{v}\in V .
\end{align}
Multiplying on the left by $\psi_{t+1}$, on the right by $\psi_{t-1}^{-1}$, and using
$\psi_t = e^{v_t}\psi_{t-1}$, this is equivalent to
\begin{align}\label{eq:constvel}
    e^{v_{t+1}}\,\mathsf{v} = \mathsf{v}\,e^{v_t} \qquad \forall\, \mathsf{v}\in V .
\end{align}
Taking $\mathsf{v} = e$ gives $e^{v_{t+1}} = e^{v_t}$, so the velocity must be constant in
time; the remaining condition $e^{v_t}\mathsf{v} = \mathsf{v}\,e^{v_t}$ is then automatic,
since $e^{v_t} \in V$ by assumption. Hence
\eqref{eq:constvel} holds precisely when the velocity is constant, that is, when $\psi$ is a
flow. For global transformations the condition holds trivially. If $\gamma \in \cG_{glob}$ then
$\gamma_t = \gamma$ for all $t$, so the transport conjugation
$\gamma_t^{-1}\mathsf{v}\,\gamma_{t-1} = \gamma^{-1}\mathsf{v}\,\gamma$ does not depend on $t$
and the induction closes. The \FE{} is therefore also $\cG_{glob}$-equivariant.

\subsection{Equivariance of the Velocity model}\label{velocity_model_equivariance}
Suppose first that $u_t[\psi \triangleright \CX_{\le t}] = \psi_tu_t[\CX_{\le t}]\psi_t^{-1}+\dot\psi_t\psi_t^{-1}$, and define $\widetilde T_{t,t_0}=\psi_tT_{t,t_0}[\CX_{\le t}]\psi_{t_0}^{-1}$. Then
\begin{align*}
\frac{d}{dt}\widetilde T_{t,t_0}
&=\dot\psi_tT_{t,t_0}\psi_{t_0}^{-1}
+\psi_t\frac{d}{dt}T_{t,t_0}\psi_{t_0}^{-1} \\
&=\left(\dot\psi_t\psi_t^{-1}
+\psi_tu_t[\CX_{\le t}]\psi_t^{-1}\right)\widetilde T_{t,t_0}
=u_t[\psi\triangleright\CX_{\le t}]\,\widetilde T_{t,t_0}.
\end{align*}
Since $\widetilde T_{t_0,t_0}=\psi_{t_0}e\psi_{t_0}^{-1}=e$, uniqueness of solutions to the ODE implies
$T_{t,t_0}[\psi\triangleright\CX_{\le t}]
=\psi_tT_{t,t_0}[\CX_{\le t}]\psi_{t_0}^{-1}$.
Conversely, suppose
$T_{t,t_0}[\psi\triangleright\CX_{\le t}]
=\psi_tT_{t,t_0}[\CX_{\le t}]\psi_{t_0}^{-1}$.
Differentiating both sides and using $\frac{d}{dt}T_{t,t_0}=u_tT_{t,t_0}$ gives
\begin{align*}
u_t[\psi\triangleright\CX_{\le t}]\,T_{t,t_0}[\psi\triangleright\CX_{\le t}]
=
\left(\dot\psi_t\psi_t^{-1}
+\psi_tu_t[\CX_{\le t}]\psi_t^{-1}\right)
T_{t,t_0}[\psi\triangleright\CX_{\le t}],
\end{align*}
and multiplying on the right by $T_{t,t_0}^{-1}[\psi\triangleright\CX_{\le t}]$ yields
$u_t[\psi\triangleright\CX_{\le t}]
=\psi_tu_t[\CX_{\le t}]\psi_t^{-1}
+\dot\psi_t\psi_t^{-1}$.

\subsection{Motion equivariance of the Velocity Based \ME{}}\label{transporter_based_FEConvLSTM_equivariance}
Here, in this section we prove Theorem~\ref{thm:motion_equivariance} which states that the transporter based \ME{} is motion equivariant.
\begin{proof}
The proof proceeds by induction on $t$. Similar to the flow-equivariance proof of \FE{}, the base case $t=0$ is trivial since, by assumption, the initial state variables $\CH_0$, $\CC_0$ and the empty sequence $\CX_{<0}$ are invariant under the action of motion transformations which establishes the equivariance at time $t=0$.
Then, we assume that the state variables $\CH_t$ and $\CC_t$ are motion equivariant at time $t$ and we show that they are also motion equivariant at time $t+1$.
For all the gate variables $z \in \{i,f,o,c\}$ we have
\begin{align}
    z_t[\psi \triangleright \CX_{\le t}](x) &= a_z\Big[ T_t[\psi \triangleright \CX_{\le t}] \cdot \left[W_{z} \star {\cal H}_t[\psi \triangleright \CX_{<t}]\right]\!(x) + [U_{z} \star (\psi \triangleright {\cal X}_t)](x)
    \Big] \\
    &= a_z\Big[  \psi_{t} T_t[\CX_{\le t}]  {\psi_{t-1}}^{-1} {\psi_{t-1}} \cdot \left[W_{z} \star {\cal H}_t[\CX_{<t}]\right]\!(x) + \psi_t \cdot [U_{z} \star {\cal X}_t](x)\Big] \\
    &=   \psi_{t}  \cdot a_z\Big[ T_t[\CX_{\le t}]\cdot [W_{z} \star {\cal H}_t[\CX_{<t}]](x) + [U_{z} \star {\cal X}_t](x)\Big] \\
    &= \psi_t \cdot z_t[\CX_{\le t}](x)\\
    & = \psi \triangleright z_t[\CX_{\le t}](x)
\end{align}
where in the second line we used the induction hypothesis for the hidden state variable together with the equivariance of the convolution under regular action of the group and also the equivariance of the transporter $T_t$ under the motion transformation. In the third line we used associativity of the group action and the fact that the nonlinearity $a_z$ is pointwise and is equivariant under the action of group elements. In the fourth line we used the definition of the gate variables.
For the cell state we have
\begin{align}
    \CC_{t+1}[\psi \triangleright \CX_{\le t}](x) &= T_t[\psi \triangleright \CX_{\le t}] \cdot {\cal C}_t[\psi \triangleright \CX_{< t}](x)  \odot f_t[\psi \triangleright \CX_{\le t}](x)   + i_t[\psi \triangleright \CX_{\le t}](x) \odot c_t[\psi \triangleright \CX_{\le t}](x)  \\
    &=  \psi_{t} T_t[\CX_{\le t}]  \psi_{t-1}^{-1} \psi_{t-1} \cdot {\cal C}_t[\CX_{< t}](x)  \odot  \psi_t \cdot f_t[\CX_{\le t}](x)
    \nonumber \\
    &\hspace{6cm}
    +\psi_t \cdot i_t[\CX_{\le t}](x) \odot \psi_t \cdot c_t[\CX_{\le t}](x) \\
        & = \psi_t \cdot \Big[T_t[\CX_{\le t}]\cdot {\cal C}_t[\CX_{< t}] \odot f_{t}[\CX_{\le t}]  +  i_t[\CX_{\le t}] \odot c_t[\CX_{\le t}] \Big](x)\\
        & = \psi_t \cdot \CC_{t+1}[\CX_{\le t}](x)\\
    & = \psi \triangleright \CC_{t+1}[\CX_{\le t}](x)
\end{align}
where in the second line we used the induction hypothesis for the cell state variable at time $t$, the equivariance of the transporter $T_t$ together with the equivariance of the gate variables which we just proved. In the third line, we factored out $\psi_t$ using the fact that the regular action distributes over the Hadamard product. Finally for the hidden state we have
\begin{align}
    \CH_{t+1}[\psi \triangleright \CX_{\le t}](x)&= o_{t}[\psi \triangleright \CX_{\le t}](x) \odot \tanh (\CC_{t+1}[\psi \triangleright \CX_{\le t}](x) ) \\
    &= \psi_t \cdot o_{t}[\CX_{\le t}](x) \odot \tanh (\psi_t \cdot \CC_{t+1}[\CX_{\le t}](x) )\\
    &= \psi_t \cdot \Big[o_{t}[\CX_{\le t}](x) \odot \tanh (\CC_{t+1}[\CX_{\le t}](x) )\Big] \\
    &= \psi_t \cdot \CH_{t+1}[\CX_{\le t}](x)\\
    & = \psi \triangleright \CH_{t+1}[\CX_{\le t}](x)
\end{align}
where in the second line we used the motion equivariance of the output gate variable and cell state variable which we also proved. In the third line we used the fact that $\tanh$ is a pointwise nonlinearity and is equivariant under the action of group elements. We also factored out $\psi_t$ using the fact that the regular action distributes over the Hadamard product. We then just replace the definition of the hidden state variable at time $t+1$. This completes the inductive step and hence the proof of motion equivariance for the \ME{} model.
\end{proof}

\subsection{Time-dependent data and motion equivariance as a gauge theory}\label{app:gauge_theory}
In this section, we provide a geometric interpretation of time-dependent data and motion equivariance in terms of principal bundles and gauge connections. This perspective naturally identifies time-dependent data as sections of an associated bundle, motions as gauge transformations, and the velocity model as a gauge connection that defines parallel transport along time through the transporter. Motion equivariance of the model can then be understood as equivariance under gauge transformations.

\paragraph{Temporal data as sections of an associated bundle.}
As our base manifold is just $\R$, our bundles are trivial and we can work with global trivializations.
We define the trivial principal $G$-bundle
\begin{align}
    P = \mathbb{R} \times G, \qquad \pi_P(t,g) = t,
\end{align}
with right action $ (t,g)\cdot h = (t, gh)$ for $h \in G$.
From this, we construct the associated bundle with fiber $F$ which is the feature space of our model:
\begin{align}
    \mathcal{E} = P \times_G F
    :=
    (P \times F) / \sim,
\end{align}
where the equivalence relation is $ (p \cdot g,f) \sim (p, g \cdot f)$ for $g \in G$, $p \in P$ and $f \in F$. Here our associated bundle is also trivial and can be identified with $\mathbb{R} \times F$.
A time-dependent signal $\mathcal{X}: \mathbb{R} \to F$ is then naturally identified with a section of this associated bundle $s_{\mathcal{X}}(t) = [(t,e),\mathcal{X}_t] \in \mathcal{E}$ where $e$ is the identity in $G$. This means that the space of time-dependent signals $\cal F$ can be identified with the space of sections of the associated bundle $\Gamma(\mathcal{E})$.

\paragraph{Gauge transformations.}
A gauge transformation of the principal bundle $P$ is a bundle automorphism commuting with the right $G$-action and preserving the base space $\R$. In the trivialization $P = \R \times G$, every gauge transformation is uniquely determined by a smooth map
$\gamma : \R \to G$,
acting fiberwise as $(t,h) \mapsto (t, \gamma_t h)$.
Therefore, the group of gauge transformations is naturally identified with the group of smooth $G$-valued functions on time which in our case is ${\cal G}_{mot}\rtimes{\cal G}_{glob}$ where ${\cal G}_{mot}$ consists of smooth motions initialized at the identity and ${\cal G}_{glob} \cong G$ is the subgroup of time-independent transformations.

A gauge transformation induces an action on the associated bundle $\mathcal{E}$ and thus on sections of $\mathcal{E}$, which, in our trivial bundle case, is simply given by
\begin{align}
    {(\gamma \triangleright \mathcal{X})}_t = \gamma_t \cdot \mathcal{X}_t \qquad \forall \gamma \in {\cal G}_{mot}\rtimes{\cal G}_{glob}
\end{align}
which is exactly the action of motion transformations on time-dependent signals that we defined in the main text. Therefore, motions are naturally identified with gauge transformations in this geometric framework.

\paragraph{Gauge connection and parallel transport.}
A connection on the principal bundle $P$ is a $\mathfrak g$-valued one-form on $P$ satisfying the standard equivariance and normalization properties, and it defines parallel transport along the time direction. In the global trivialization $P \cong \mathbb{R} \times G$, it is represented by a $\mathfrak{g}$-valued one-form on $\mathbb{R}$ of the form $u_t\, dt$, where $u_t \in \mathfrak{g}$.
Under a gauge transformation $\gamma : \mathbb{R} \to G$, the connection transforms as
\begin{align}
    \gamma \triangleright u_t= \operatorname{Ad}_{\gamma_t}(u_t) + \dot{\gamma}_t \gamma_t^{-1}
\end{align}
where $\operatorname{Ad}_{\gamma_t}(u_t) = \gamma_t u_t \gamma_t^{-1}$ denotes the adjoint action of $G$ on its Lie algebra $\mathfrak g$.
Given a connection $u_t$, one can define a covariant derivative and the associated parallel transport along time. The covariant derivative measures the intrinsic time variation of a section, i.e. the variation after subtracting the contribution induced by gauge transformations which in our case are motions.

The parallel transport from time $t_1$ to $t_2$ induced by the connection $u_t$ is given by the associated Wilson line, i.e., the time-ordered exponential of the connection:
\begin{align}
    T_{t_2,t_1} = \mathcal{T} \exp\left(\int_{t_1}^{t_2} u_s \, ds \right).
\end{align}
Equivalently, it is the unique solution of the differential equation $\frac{d}{dt} T_{t,t_0} = u_t T_{t,t_0}$ with initial condition $T_{t_0,t_0} = e$. Its action on a time-dependent signal (viewed as a section) transports features from time $t_1$ to time $t_2$ by parallel transport along the connection $u_t$, meaning that the transported section remains covariantly constant with respect to the induced covariant derivative.

\paragraph{Parallel transport and model dynamics.}
Given a gauge connection in the continuous-time limit, we can define a covariant time derivative and express the model dynamics in terms of this covariant derivative. This formulation yields an evolution equation that is equivariant under gauge transformations, and hence under motions as well.
In the discrete-time setting, we instead use the parallel transporter induced by the connection to modify the update equations. Concretely, the feature produced at time $t$ is transported to time $t+1$ in a parallel manner and then combined with the newly computed features at time $t+1$ to form the updated hidden state. This construction ensures that the resulting discrete-time dynamics are motion equivariant by design.
The mechanism echoes the `gauge principle,' whereby equivariance is achieved through the introduction of an additional gauge degree of freedom.

\section{Experimental Details}
\label{app:experimental_details}

\subsection{Future-Frame Prediction}
\label{app:pred}

\subsubsection{Moving MNIST Dataset}
\label{app:dataset}
We construct a synthetic dataset of $n_d$ MNIST digits moving with time-dependent
velocities on the discrete torus $\mathbb T^2_{S}$, obtained by zero-padding the
$28\times28$ MNIST digits to $S\times S$ with $S \ge 28$. Each digit
$i = 1,\ldots,n_d$ carries a velocity $v^{(i)}_t \in V_{N}$ drawn from the
velocity lattice
$V_{N} = \{\, v \in \mathbb{Z}^2 \mid \|v\|_\infty \le N \,\}$.
A frame is the superposition of the individual digits,
\begin{align}
    \CX_t = \sum_{i=1}^{n_d} \CX^{(i)}_t,
    \qquad
    \CX^{(i)}_t = \psi^{(i)}_t \cdot \CX^{(i)}_0 ,
\end{align}
where $\psi^{(i)}$ is the discrete motion generated by the velocity sequence
$\{v^{(i)}_t\}$ as defined in Section~\ref{sec:dynamical_symmetries}.

\paragraph{Motions.}
We construct a stochastic generative process that produces a rich distribution of motions. To govern the temporal dynamics, we introduce two key parameters: $p_{\text{change}}$ to control the frequency of velocity updates, and $p_{\text{smooth}}$ to control the spatial smoothness of the motion. At each time step $t$, a velocity update is triggered independently with probability $p_{\text{change}}$. When an update occurs, the new velocity transitions to an immediate neighboring state on the lattice with probability $p_{\text{smooth}}$. With probability $1 - p_{\text{smooth}}$, it is instead sampled uniformly from the full lattice $V_N$ (excluding the current velocity), allowing for a long-tail of sudden, larger accelerations. Formally, the velocity evolves as:
\begin{equation}
v_t =
\begin{cases}
\mathrm{update}(v_{t-1}) & \text{w.p. } p_{\text{change}}, \\
v_{t-1} & \text{otherwise,}
\end{cases}
\qquad
\mathrm{update}(v) =
\begin{cases}
\mathrm{nbr}(v) & \text{w.p. } p_{\text{smooth}}, \\
\mathcal{U}(V_N \setminus \{v\}) & \text{w.p. } 1 - p_{\text{smooth}}.
\end{cases}
\end{equation}
where $\mathrm{nbr}(v) = v + \varepsilon$ draws $\varepsilon$ uniformly from the eight unit spatial offsets. If the proposal leaves the lattice $V_N$, the update holds ($\mathrm{nbr}(v) = v$), ensuring the stationary velocity distribution remains perfectly uniform. Alternatively, to generate trajectories with more structured durations, we also employ a piecewise-constant formulation where velocity is strictly held for a duration $\tau \sim \mathcal{U}(s_{\text{min}}, s_{\text{max}})$ before an update is triggered using the identical rules.

\paragraph{Kinematic Metrics.}
Because out-of-bounds proposals are held rather than resampled, a triggered update does not strictly guarantee a change in velocity. To accurately classify the motion regimes, we define two metrics measured from the realised trajectories. First, the \emph{realised} switching rate $r$ (the empirical probability of a velocity transition on a given frame) is strictly lower than the nominal trigger rate $p_{\text{change}}$:
\begin{equation}
r = p_{\text{change}} \bigl(1 - p_{\text{smooth}}(1 - \mathrm{acc}_N)\bigr),
\end{equation}
where $\mathrm{acc}_N$ is the lattice-dependent probability that a unit-offset stays within $V_N$. Second, to strictly decouple the frequency of these transitions from their severity, we define the jump magnitude $|\Delta v|$ as the $L_\infty$ norm of the velocity difference, $\max(|v_{x,t} - v_{x,t-1}|, |v_{y,t} - v_{y,t-1}|)$, averaged strictly over the subset of frames where a transition actually occurred. By conditioning the expectation on $v_t \neq v_{t-1}$, the jump magnitude isolates the purely spatial severity of the transitions independently of their frequency.

\paragraph{Velocity Freezing and Model Evaluation.}
As outlined in the main text, every sequence sets \texttt{freeze\_after} equal to the context length ($T_{context}$). This yields a two-phase trajectory: a time-varying phase during the encoding context, followed by a constant-velocity phase over the entire predicted rollout. Neither model is given access to the ground truth velocities; they must infer them solely from the video frames. By freezing the rollout velocity, we establish a fair, perfectly balanced competition between the two architectures:
\begin{enumerate}
\item \textbf{\ME{} (Explicit Estimation):} To succeed, \ME{} must learn a robust encoding of the digits \emph{and} successfully estimate the correct boundary velocity (via phase correlation) in order to explicitly transport those representations into the future.
\item \textbf{\FE{} (Implicit Routing):} \FE{} does not estimate a single velocity vector, but instead maintains separate spatial channels for all valid velocities. To succeed, it must learn a robust encoding of the digits \emph{and} successfully route those representations into the correct velocity channels by the end of the context.
\end{enumerate}
Because the rollout velocity is constant, it is theoretically observable from the final context frames for both models. By strictly freezing the kinematics during the rollout, the evaluation isolates their ability to encode and assign velocities, decoupling the competition from the impossible task of guessing unpredictable future velocity changes.

\subsubsection{Training and Evaluation Setup}
\label{app:training_setup}

\paragraph{Dataset Configuration and Splits.}
Our standard training configuration utilizes the piecewise motion regime on $\mathbb{T}^2_{36}$ with $\tau \sim \mathcal{U}(3, 6)$, $p_{\text{smooth}} = 0.80$, and $V_2$ (where $\mathrm{acc}_2 = 2/3$). The expected segment duration of 4.5 frames implies a nominal switching rate of $2/9 \approx 0.222$. Applying the realized rate correction, the training dataset operates at a realized switching rate of $r = 0.163$ changes per frame, with a mean realized jump magnitude of $|\Delta v| = 1.40$. We use $54{,}000/6{,}000$ training/validation sequences (a $90/10$ split of the MNIST training set) and $10{,}000$ test sequences (the MNIST test set), ensuring digit identity never crosses between training and evaluation. Training and validation sequences are generated on the fly: fresh digit pairs, placements, and velocity schedules are drawn on every access so no sequence is seen twice. The test set and every evaluation benchmark are instead pre-generated from fixed seeds.

\paragraph{Optimization and Hyperparameters.}
All three models are trained identically: $25$-frame sequences ($15$ context, $10$ predicted), optimized via MSE${}+{}$L1 loss computed exclusively on the predicted frames. We use the Adam optimizer with an initial learning rate of $10^{-3}$ and \texttt{ReduceLROnPlateau} scheduling (factor $0.5$, patience $5$, floor $10^{-6}$), along with gradient clipping at $1.0$. Training proceeds with a batch size of $32$ for $50$ epochs (totaling $84{,}400$ optimizer steps) on a single NVIDIA A100 80\,GB GPU. Early stopping was never triggered.

\paragraph{Parameter Budget.}
All three models have exactly $47{,}681$ parameters, all trainable: $38{,}144$ in the
recurrent cell and $9{,}537$ in the decoder. The counts agree exactly rather than
approximately, because weights are shared across \FE{}'s $25$ velocity channels and
\ME{}'s $K=2$ slots, and because \ME{}'s velocity model contributes no parameters at all
(Appendix~\ref{app:velocity_model}). The three architectures therefore differ only in how
many hidden-state copies they carry per step: $1$, $25$ and $2$ respectively, and
not in their number of learnable parameters.

\paragraph{Validation Tracking.}
Validation loss is tracked at two resolutions: a fixed $256$-sequence probe scored every $25$ batches, which resolves training dynamics without resampling noise, and the full $6{,}000$-sequence validation split scored once per epoch. The two curves agree closely throughout training, confirming that the high-frequency probe is unbiased.

\subsubsection{Length and Motion Generalization}
\label{app:length_motion_generalization}
\begin{figure*}[t]
    \centering
    \bleedgraphics{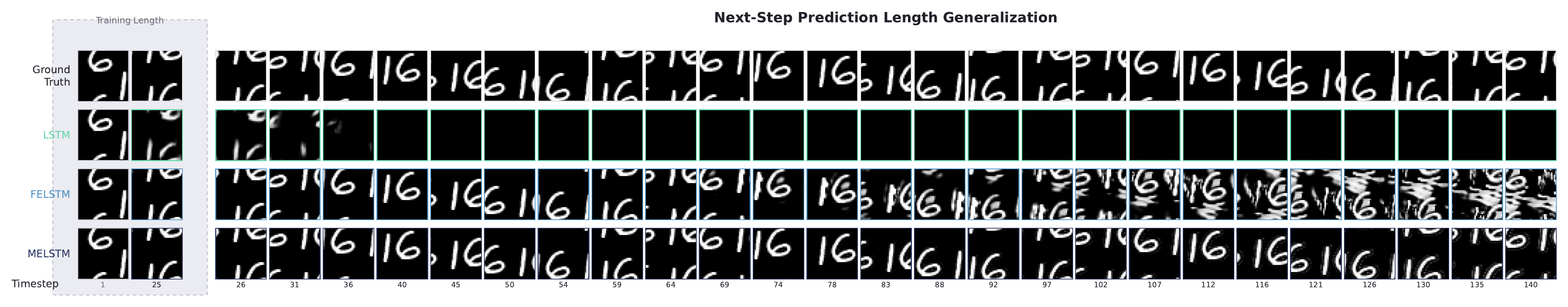}\\[-1ex]
    \bleedgraphics{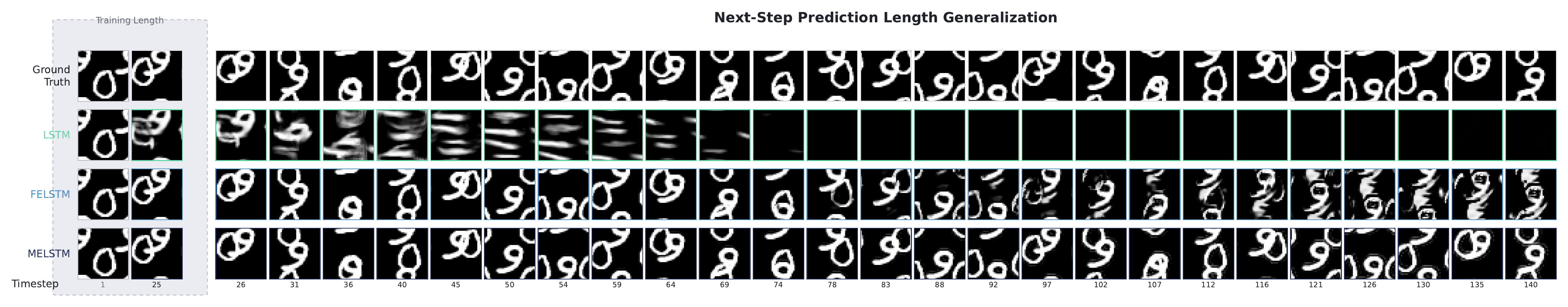}
    \caption{\textbf{Length generalization beyond the trained horizon.} Two
    held-out sequences; rows are the ground truth, \CL{}, \FE{}, and \ME{}. The
    dashed box marks the $25$ frames seen in training ($15$ context $+$ $10$
    predicted); the rest is extrapolation, run out to $t=140$, past the
    $85$-step horizon scored in Figure~\ref{fig:len_gen_boundary_age}. \CL{}
    collapses almost immediately, \FE{} smears into high-frequency artifacts
    late in the rollout, and \ME{} stays sharp throughout.}
    \label{fig:rollout_seq0}
    \label{fig:rollout_seq12}
\end{figure*}

\paragraph{Length Generalization.}
We extend the 10-frame trained rollout to an 85-frame free-running rollout ($8.5\times$ the trained horizon), predicting $t=16$ through $t=100$ after $15$ context frames; Figure~\ref{fig:len_gen_boundary_age} reports the resulting rollout MSE. Because velocity is frozen throughout (the velocity active at the last context transition is held constant), the test isolates stable autoregressive spatial extrapolation from the ability to predict future velocity changes. Figure~\ref{fig:rollout_seq0} shows the same behaviour qualitatively, run past the scored horizon to $t=140$.

\begin{figure}[!t]
\centering
{\small\setlength{\tabcolsep}{6pt}
\begin{tabular}{llcccc}
\toprule
Regime & Kinematics & $r$ & $|\Delta v|$ & $\rho$ & net disp (px) \\
\midrule
\textbf{trained} & Baseline training motion & $\mathbf{0.163}$ & $\mathbf{1.40}$ & $\mathbf{0.121}$ & $\mathbf{23.2}$ \\
\midrule
\texttt{constant} & Pure linear flow & $0.000$ & --- & $0.000$ & $27.5$ \\
\texttt{rare\_teleport} & Rare, violent jumps & $0.050$ & $2.41$ & $0.063$ & $23.7$ \\
\texttt{regular\_fast} & Fast, regular steps & $0.344$ & $1.40$ & $0.251$ & $19.1$ \\
\texttt{random\_walk} & Constant tiny jitters & $0.657$ & $1.00$ & $0.361$ & $21.7$ \\
\texttt{churn\_mid} & Frequent large jumps & $0.451$ & $1.99$ & $0.468$ & $13.5$ \\
\texttt{churn\_half} & Frequent violent jumps & $0.498$ & $2.44$ & $0.629$ & $10.9$ \\
\texttt{churn\_max} & Pure erratic chaos & $0.735$ & $2.42$ & $0.927$ & $8.5$ \\
\bottomrule
\end{tabular}}

\vspace{1.1em}
\includegraphics[width=0.55\linewidth]{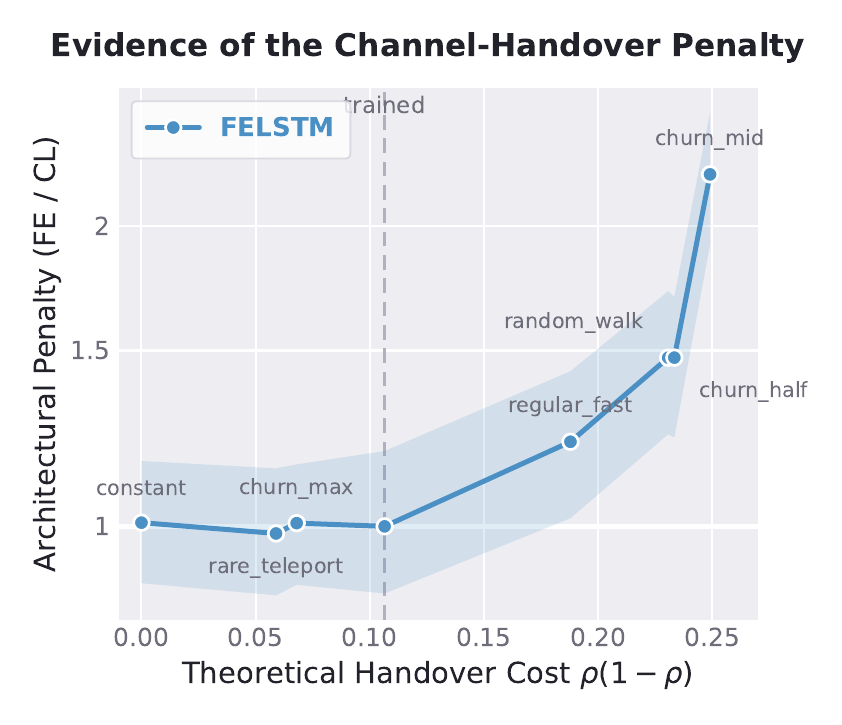}
\caption{\textbf{The channel-handover penalty across the seven motion regimes.}
\emph{Top:} kinematics of the seven evaluation regimes and the training
baseline. The unpredictability $\rho$ collapses the switching rate $r$ and the
jump magnitude $|\Delta v|$ into a single number; net context displacement falls
as $\rho$ rises, because the trajectories become diffusive.
\emph{Bottom:} \FE{}'s architectural penalty (MSE normalised by the \CL{}
baseline) against $\rho(1-\rho)$. Because \FE{} allocates separate channels to
discrete velocities, a velocity change forces a handover that strands
accumulated temporal history; we model its cost as the product of the
disruption ($\rho$) and the predictive value of what is stranded ($1-\rho$).
The penalty is negligible up to $\rho(1-\rho)\approx0.1$, including under pure
chaos (\texttt{churn\_max}, $\rho\to1$), where the stranded history is
worthless, and grows steeply above it. \texttt{random\_walk} (frequent tiny
jitters) and \texttt{churn\_half} (less frequent but far larger jumps) are
kinematically opposite, yet sit at nearly symmetric points of the parabola and
show nearly equal penalties.}
\label{fig:handover-penalty}
\end{figure}

\paragraph{Motion Generalization.}
To test robustness to changes in motion, we rescore the trained checkpoints,
without retraining, on test sequences whose context motion differs from training;
the velocity is still frozen at the context boundary. Each model is scored by its
median per-sequence MSE over the trained horizon ($t=16$--$25$), divided by its
own median on a reference motion with the training switching rate and jump sizes
(marked ``trained'' in Figure~\ref{fig:motion_generalization}), so $1$ means the
new motion costs the model nothing. This removes the differences in overall error
between the models, but not the difficulty of the new motion itself, which every
model faces: \CL{}'s curve reflects that difficulty, and the ratio of \FE{} to
\CL{} (Figure~\ref{fig:handover-penalty}) isolates the cost specific to the \FE{} architecture.

\paragraph{The Channel-Handover Penalty.}
In the main text, we ordered the motion regimes by their kinematic unpredictability ($\rho$). To quantify this metric, let $T_d$ denote circular translation by $d$ and $m(d)=\tfrac{1}{HW}\,\mathbb E_x\|x-T_dx\|^2$ the mean squared image difference induced by displacement $d$. We define:
\begin{equation}
\rho \;=\;\frac{\mathbb E\bigl[m(\Delta v)\bigr]}
                {\mathbb E\bigl[m(v_t)\bigr]},
\label{eq:rho}
\end{equation}
where the numerator is the one-step error of persisting the previous velocity, and the denominator is the error of assuming no motion. Thus, $\rho=0$ indicates pure flow, while $\rho \to 1$ indicates the past velocity offers no predictive advantage.

This definition makes the product described in the main text concrete.
Handovers happen at the switching rate $r$, and $\rho$ also weights each one by
the size of its jump, so it measures how disruptive the handovers are, while the
predictive value of the history abandoned at a handover is roughly proportional
to $1-\rho$. We therefore use $\rho(1-\rho)$ as a simple model of the \FE{}
penalty. It is a heuristic rather than a derived law, and the relation is not
linear: the penalty stays near $1$ for the four regimes with
$\rho(1-\rho)\le0.11$ (\texttt{constant}, \texttt{rare\_teleport},
\texttt{churn\_max} and the training motion) and then grows through
\texttt{regular\_fast}, \texttt{random\_walk} and \texttt{churn\_half} to
\texttt{churn\_mid}, the regime closest to $\rho=0.5$
(Figure~\ref{fig:handover-penalty}).

\subsubsection{Velocity Model}\label{app:velocity_model}

\begin{figure*}[t]
    \centering
    \bleedgraphics{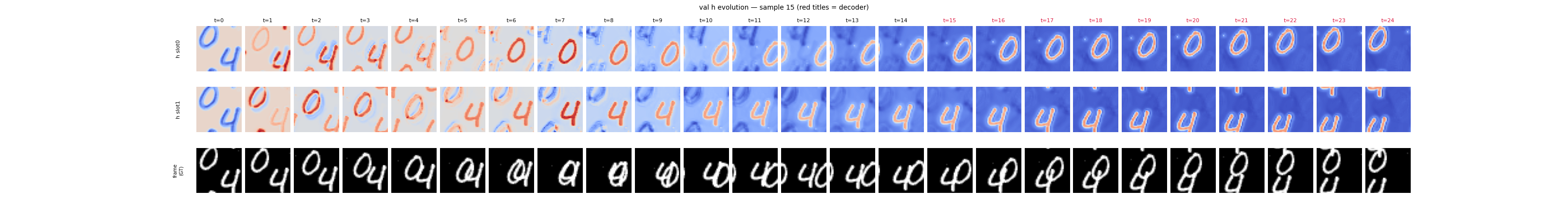}\\[-1ex]
    \bleedgraphics{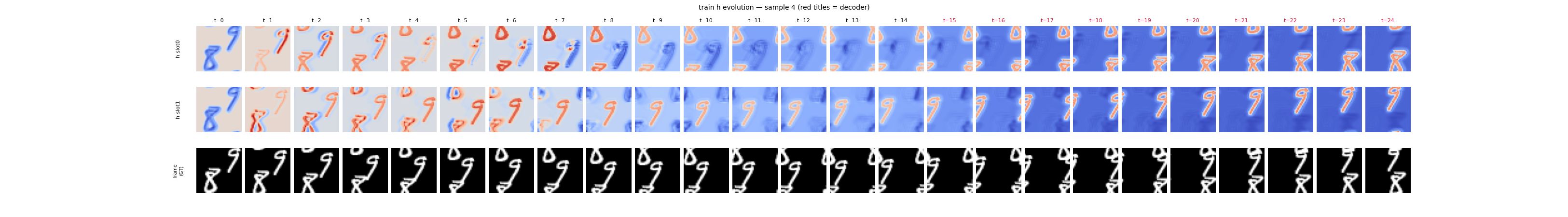}
    \caption{\textbf{Slot specialisation.} Hidden-state evolution for a
    validation sequence (top) and a training sequence (bottom). Within each
    block the first two rows show the channel-averaged states
    $\bar \CH^1, \bar \CH^2$ of the $K=2$ slots and the third the corresponding
    frame; column titles give $t$, in red once the decoder takes over. Each slot
    localises onto one digit and tracks it, and this is the template that
    Eq.~\eqref{eq:track} correlates against the incoming frame.}
    \label{fig:val_MELSTM_2slots}
    \label{fig:train_MELSTM_2slots}
\end{figure*}

As introduced in the main text, we estimate local velocities using a slot tracking
mechanism. Specifically, each slot's velocity is computed via phase correlation between its
own hidden state and the current frame.

\paragraph{Phase correlation.}
Fourier phase correlation is a global frequency-domain technique for estimating the relative
translation between two images or signals. Given a reference image $\CX_1$ and a shifted
image $\CX_2$, the normalized cross-power spectrum is computed as
\begin{equation}
R \;=\; \frac{\mathfrak{F}[\CX_1]\odot\overline{\mathfrak{F}[\CX_2]}}
             {\bigl|\mathfrak{F}[\CX_1]\odot\overline{\mathfrak{F}[\CX_2]}\bigr|},
\end{equation}
where $\mathfrak{F}$ denotes the two-dimensional Fourier transform, $\odot$ is the Hadamard
product, and the overline represents complex conjugation. Applying the inverse Fourier
transform to $R$ yields a correlation surface whose peak indicates the estimated translation
between the two images:
\begin{equation}
    u^* = \argmax_{u} \mathfrak{F}^{-1}[R](u).
\end{equation}
The height of this peak serves as a confidence measure for the estimate. We denote this
entire procedure as
\[ u^* = \mathcal{P}_1(\CX_1, \CX_2), \]
where $\mathcal{P}_1$ is the phase correlation operator returning the translation vector that
corresponds to the dominant peak of the correlation surface. The subscript indicates that only
the single strongest peak is considered.

\paragraph{From a local problem to a global one.}
A frame containing $N$ independently moving objects does not possess a single uniform velocity.
Consequently, recovering motion strictly from consecutive frames $(X_{t-1}, X_t)$ is
inherently a \emph{local} problem: the frame must be segmented and the correspondence resolved
region by region, which is the primary challenge addressed by optical flow models. By contrast,
a global estimator like phase correlation is better suited for determining a single global
motion. Applying it directly to raw frames and extracting the $K$ strongest peaks via
$\mathcal{P}_K(X_{t-1}, X_t)$ does yield $K$ velocities at each timestep $t$. However, these
velocities form an unordered set; there is no inherent mechanism to associate a peak at time
$t$ with its corresponding peak at time $t-1$. Converting these sets into coherent trajectories
requires a separate assignment step at every timestep, typically relying on continuity
heuristics or matching algorithms, where any error becomes unrecoverable.

The introduction of the hidden state circumvents these difficulties. For $t \ge 2$, each slot
correlates its own channel-averaged hidden state against the incoming frame:
\begin{equation}
u^k_t \;=\; \mathcal{P}_1\bigl(\bar \CH^k_{t},\, \CX_t\bigr),
\qquad
\bar \CH^k = \tfrac{1}{C_h}\textstyle\sum_c \CH^k[c].
\label{eq:track}
\end{equation}
Through this operation, slot $k$ effectively queries where \emph{its specific} content has
moved. Once a slot encapsulates a single coherently moving object, this query has exactly one
valid answer: a single global displacement between the two fields, which phase correlation
computes in closed form. In this framework, the local tracking problem is not merely solved,
but rather absorbed: the recurrent dynamics resolve the segmentation within the state itself,
allowing the estimator to face only a global displacement problem. Furthermore, each computed
velocity is directly bound to the slot that requested it. Consequently, the slot's identity
inherently acts as the tracking mechanism, naturally dissolving the correspondence problem
without requiring explicit matching.

\paragraph{Bootstrap.}
At $t = 0$, all slot states are initialized to zero. Since warping a zero state has no effect
and every slot receives identical input, the $K$ slots would ordinarily remain identical
copies of one another indefinitely. Breaking this symmetry requires an initial velocity
estimate that does not originate from the hidden state. We resolve this by computing
$u^{1..K} = \mathcal{P}_K(X_0, X_1)$, which seeds the slots with the $K$ strongest peaks from
the first frame pair. This initial step is the only instance where correlation is performed
directly between raw frames, and the only time more than one peak is extracted. For every
timestep $t \ge 2$ onward, the velocities are individually determined via
Eq.~\eqref{eq:track}.

\begin{figure}[t]
    \centering
    \includegraphics[width=0.55\linewidth]{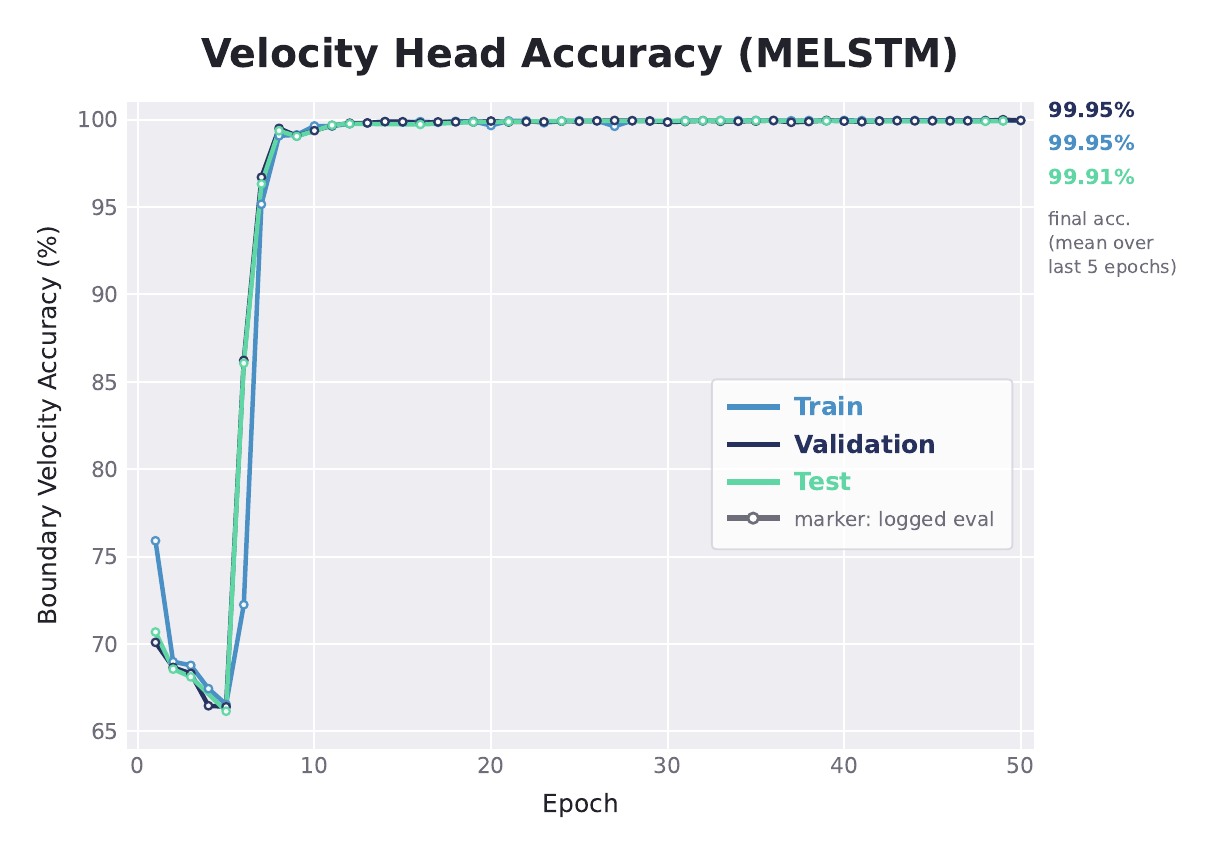}
\caption{\textbf{Velocity accuracy} (one training run). The estimator is
parameter-free and receives no gradient, so the rise from $\approx 66\%$ at epoch
$5$ to above $99.5\%$ by epoch $8$ reflects the hidden-state template improving,
not the estimator being fitted. Final values are means over the last five epochs.
In the same run, the validation loss stays on a plateau near $7\times10^{-2}$
until about epoch $6$ and then falls by an order of magnitude while the accuracy
climbs: the model leaves the plateau when the slots lock onto their digits. The
five seeds of Figure~\ref{fig:trainingeff} leave it at different times, which
smooths their mean.}
    \label{fig:velcity_accuracy}
\end{figure}

\paragraph{The state and the estimate improve together.}
Because the phase correlation estimator contains no trainable parameters, the observed
velocity accuracy in Figure~\ref{fig:velcity_accuracy}---which rises from $\approx 66\%$ to
over $99.9\%$---indicates that the underlying \emph{template} is improving. Specifically, the
hidden state progressively evolves into a cleaner, spatially localized representation of the
object that each slot tracks (as visualized in Figure~\ref{fig:val_MELSTM_2slots}). These two
quantities are coupled in a mutually reinforcing loop: a more accurate velocity transports the
state more faithfully, a better-transported state forms a cleaner template, and a cleaner
template subsequently yields a sharper velocity estimate. No external mechanism closes this
loop, and no explicit motion supervision is provided. Instead, the prediction loss shapes the
hidden state $h$, and the estimator directly consumes this increasingly refined state.
Therefore, the upward trend in accuracy serves as a direct probe of the model's representation
quality, rather than a learning curve for the estimator itself.

\subsubsection{Training with Frame-Pair Velocities}
\label{app:pred_xtrain}

\begin{figure}[t]
\centering
\begin{minipage}[c]{0.75\linewidth}
  \centering
  \includegraphics[width=\linewidth]{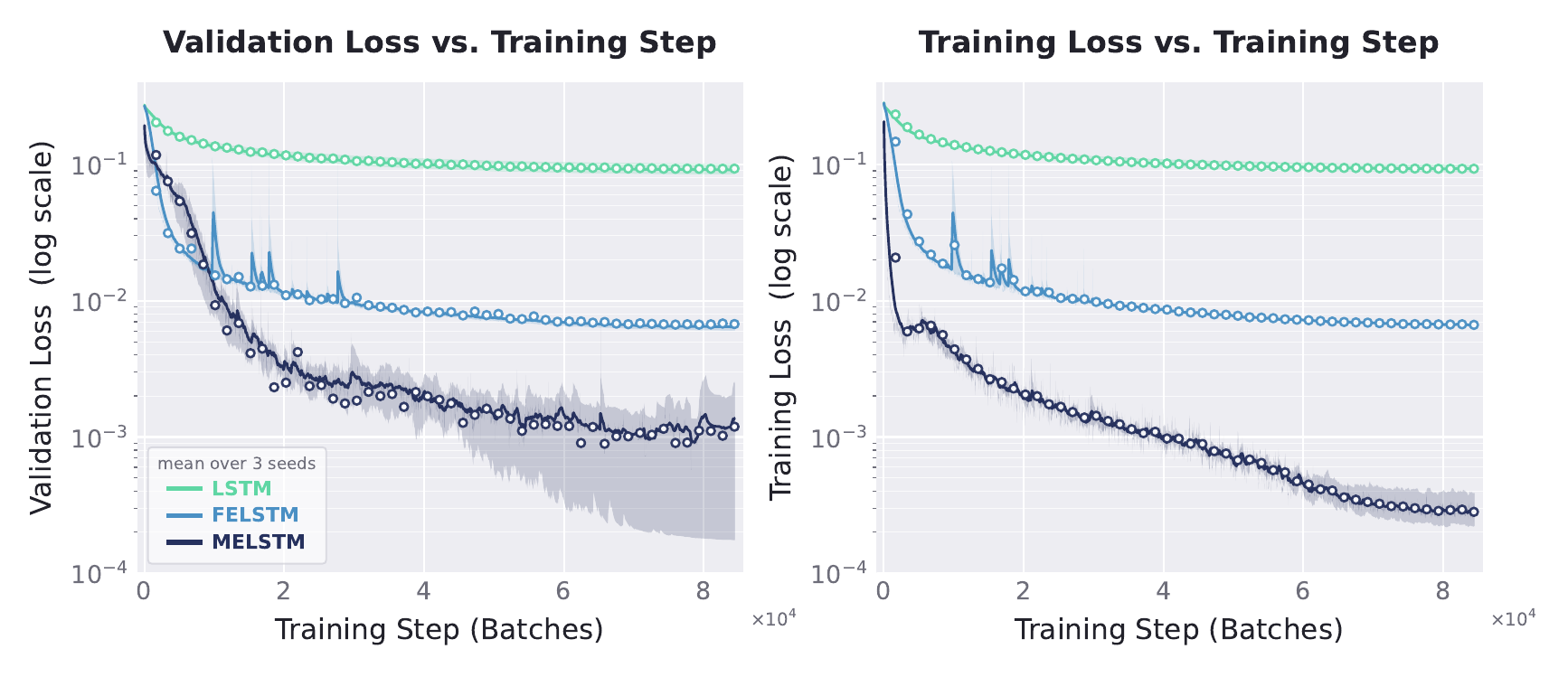}
\end{minipage}\hfill
\begin{minipage}[c]{0.24\linewidth}
  \centering
  \footnotesize
  \setlength{\tabcolsep}{2.5pt}
  \renewcommand{\arraystretch}{1.35}
  \begin{tabular}{lc}
  \toprule
  Model & \begin{tabular}{@{}c@{}}MSE $\downarrow$\\[-1pt]{\tiny test}\\[-3pt]{\tiny $t=100$}\end{tabular} \\
  \midrule
  LSTM   & \begin{tabular}{@{}c@{}}$3.01_{\pm .27}$e$-$2\\[-2pt]$1.38_{\pm .06}$e$-$1\end{tabular} \\
  \addlinespace[2pt]
  FELSTM & \begin{tabular}{@{}c@{}}$1.53_{\pm .02}$e$-$3\\[-2pt]$4.22_{\pm 6.16}$e$-$1\end{tabular} \\
  \addlinespace[2pt]
  MELSTM & \begin{tabular}{@{}c@{}}$\bm{3.17_{\pm 2.68}}$\textbf{e}$\bm{-}$\textbf{4}\\[-2pt]$\bm{5.34_{\pm 2.61}}$\textbf{e}$\bm{-}$\textbf{4}\end{tabular} \\
  \bottomrule
  \end{tabular}
\end{minipage}
\caption{\textbf{Stochastic frame-pair velocities.} In training, \ME{} uses
frame-pair velocities with a probability that decays to $0$ by epoch $25$;
validation uses slot tracking only. Losses (MSE${}+{}$L1) are means over three
seeds with the min--max range shaded; faint traces are recorded every $25$
batches and circles once per epoch (Appendix~\ref{app:training_setup}). Table:
test MSE over the trained horizon and at $t=100$; mean $\pm$ s.d.}
\label{fig:xstocos}
\end{figure}

\begin{figure}[!b]
\centering
\begin{minipage}[c]{0.75\linewidth}
  \centering
  \includegraphics[width=\linewidth]{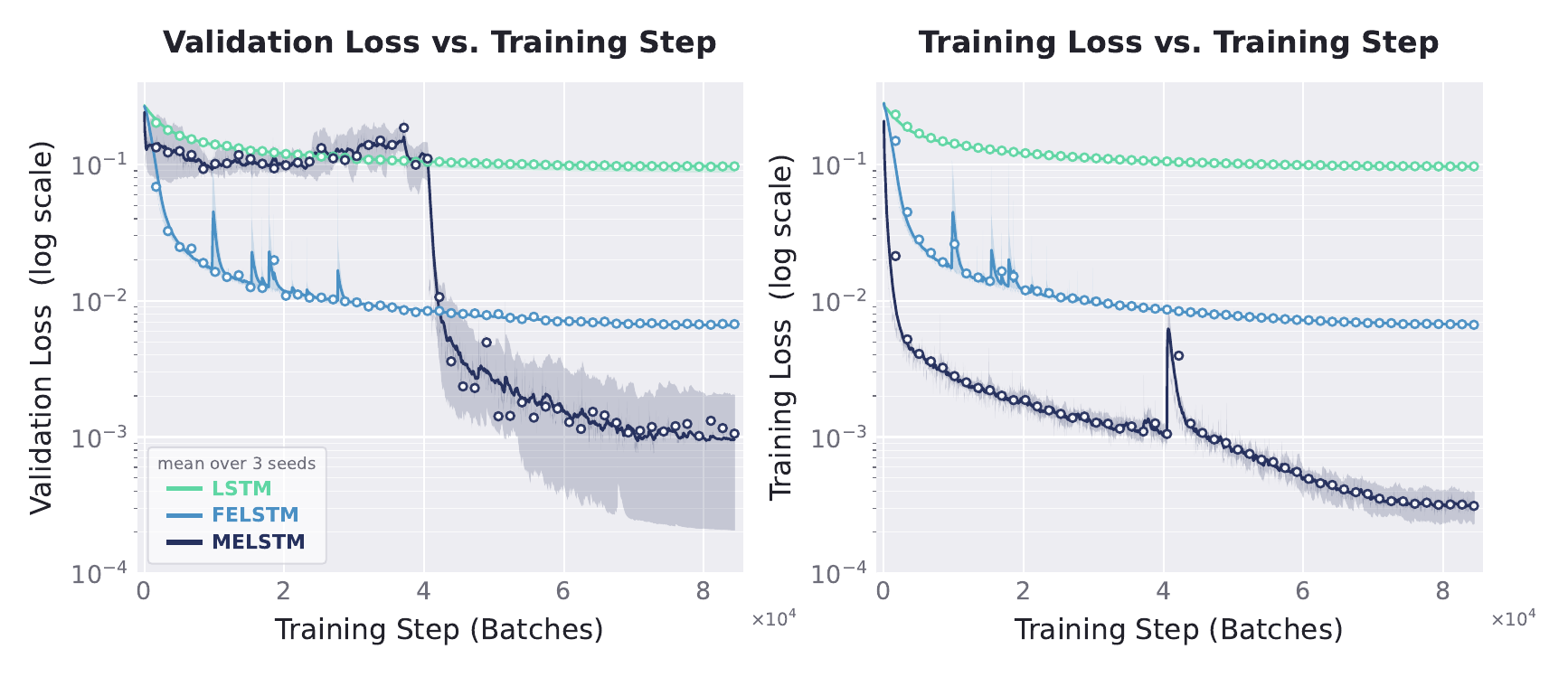}
\end{minipage}\hfill
\begin{minipage}[c]{0.24\linewidth}
  \centering
  \footnotesize
  \setlength{\tabcolsep}{2.5pt}
  \renewcommand{\arraystretch}{1.35}
  \begin{tabular}{lc}
  \toprule
  Model & \begin{tabular}{@{}c@{}}MSE $\downarrow$\\[-1pt]{\tiny test}\\[-3pt]{\tiny $t=100$}\end{tabular} \\
  \midrule
  LSTM   & \begin{tabular}{@{}c@{}}$3.15_{\pm .21}$e$-$2\\[-2pt]$1.36_{\pm .03}$e$-$1\end{tabular} \\
  \addlinespace[2pt]
  FELSTM & \begin{tabular}{@{}c@{}}$1.53_{\pm .03}$e$-$3\\[-2pt]$4.35_{\pm 6.06}$e$-$1\end{tabular} \\
  \addlinespace[2pt]
  MELSTM & \begin{tabular}{@{}c@{}}$\bm{4.28_{\pm 4.74}}$\textbf{e}$\bm{-}$\textbf{4}\\[-2pt]$\bm{7.59_{\pm 9.52}}$\textbf{e}$\bm{-}$\textbf{4}\end{tabular} \\
  \bottomrule
  \end{tabular}
\end{minipage}
\caption{\textbf{Frame-pair curriculum.} As Figure~\ref{fig:xstocos}, but
training uses frame-pair velocities on the first steps of the context, a window
that shrinks from the whole context to none by epoch $25$.}
\label{fig:xcur}
\end{figure}

Slot tracking needs a hidden state that is a good template, which takes several
epochs to form (Figure~\ref{fig:velcity_accuracy}), whereas velocities read from a
frame pair do not depend on the state. We therefore also trained \ME{} with
frame-pair velocities that are withdrawn by epoch $25$: the two strongest peaks of
$\mathcal P_2(\CX_{t-1},\CX_t)$ are matched to the two slots by closeness to their
previous velocities and replace slot tracking (Eq.~\eqref{eq:track}). In the
\emph{stochastic} schedule, each step $t\ge2$ of each training sequence uses them
with a probability that decays from $1$ to $0$ along a half cosine. In the
\emph{curriculum}, the steps $2\le t<T_x$ use them, where $T_x$ decreases linearly
from $15$, the whole context, to $2$, so the latest steps switch to slot tracking
first. From epoch $25$ on, training uses slot tracking only; validation and
testing always do.
With the stochastic schedule, every step is occasionally slot-tracked from the
second epoch on, and the validation loss falls steadily from the start
(Figure~\ref{fig:xstocos}). With the curriculum, the earliest steps switch last:
the validation loss stays near that of \CL{} until they do, at epoch $25$, and then
drops sharply, while the training loss jumps briefly (Figure~\ref{fig:xcur}). Both
schedules end at about the test error of training with slot tracking throughout
(Figure~\ref{fig:trainingeff}).

\subsection{Motion-Only MNIST Classification}
\label{app:cf}

Classification shares with the future-frame prediction experiment the $36\times36$ torus, the piecewise
motion of Appendix~\ref{app:dataset} with segments of $3$--$6$ frames on the
lattice $V_2$, the recurrent cell, and the optimizer and learning-rate schedule
of Appendix~\ref{app:training_setup}. It differs in how frames are rendered, in
the task head and in the loss.

\paragraph{Data.}
Each sequence holds one digit. Its MNIST glyph is binarised at intensity $0.3$
into a mask $\chi$, equal to one on the digit and zero elsewhere, and placed
uniformly at random on $\mathbb T^2_{36}$. The digit and the background have
independent white-noise textures $\eta^{\mathrm d}$ and $\eta^{\mathrm b}$ (unit
variance, independent across pixels) and their own motions $\psi^{\mathrm d}$
and $\psi^{\mathrm b}$, generated by velocities $v^{\mathrm d}_t$ and
$v^{\mathrm b}_t$. Frame $t$ is
\begin{equation}
    \CX_t \;=\; \psi^{\mathrm d}_t\cdot\bigl(\chi\odot\eta^{\mathrm d}\bigr)
    \;+\; \bigl(1-\psi^{\mathrm d}_t\cdot\chi\bigr)\odot
    \bigl(\psi^{\mathrm b}_t\cdot\eta^{\mathrm b}\bigr),
\label{eq:cf_frame}
\end{equation}
clipped at $\pm3$ and mapped affinely to $[0,1]$. The first term is the digit's
texture inside its moved mask; the factor $1-\psi^{\mathrm d}_t\cdot\chi$ is zero
on the digit, so the second term is the moved background with the digit cut out.
The velocities of the digit and of the background are each drawn from the
motion generator of the prediction experiment, except that smooth updates are drawn uniformly from the valid lattice neighbours, giving a realised switching rate of $0.193$ and a mean jump of $1.19$.  The $54{,}000$/$6{,}000$ training/validation glyphs are a $90/10$ split of the
MNIST training set and the $10{,}000$ test glyphs are the MNIST test set, so no
glyph is shared between training and evaluation. Training sequences are
rendered afresh on every access; validation and test sequences are fixed.
No single frame reveals the digit: binarisation discards its intensity, both
layers have the same noise statistics, and white noise has no correlation between
neighbouring pixels, so nothing marks the digit's outline.
A single-frame CNN (three circular stride-$2$ convolutions with
GroupNorm and global pooling) accordingly scores $12.2\%$ against $10\%$ chance,
so the digit can only be recovered from how its pixels move together. Smoothing the textures with a Gaussian filter would leak the outline, because
the correlation it creates breaks where the two textures meet: with a standard
deviation of $0.5$, $1$ or $2$ pixels, the same CNN reaches $13\%$, $24\%$ and
$36\%$.

\paragraph{Classifier and training.}
The head $g_\theta$ applies three $3\times3$ convolutions with stride $2$,
circular padding, $16$ channels, BatchNorm and ReLU, concatenates global average
and max pooling, and ends in an MLP $32\to32\to10$; with the recurrent cell this
gives $48{,}842$ parameters for every model. The velocity axis is max-pooled as well.
Before every evaluation, the head's BatchNorm statistics are re-estimated on
$50$ training batches with the current weights (precise
BatchNorm; \citealp{wu2021rethinking}). Training differs from the future-frame prediction experiment in the
cross-entropy loss, batch size $64$, plateau patience $4$ and $80$ epochs
($67{,}520$ steps); the checkpoint with the best validation accuracy is tested
once.

\paragraph{Velocity estimation in \ME{}.}
At $t=1$, the two peaks of $\mathcal P_2(\CX_0,\CX_1)$
(Appendix~\ref{app:velocity_model}) seed the two slots, one for the digit and one
for the background. At later steps, each new peak goes to the slot whose previous
velocity is nearest, which solves the assignment problem as long as velocities
change little from one frame to the next. With slot tracking
(Eq.~\eqref{eq:track}) instead, test accuracy is about one point lower, and the
digit also appears in the hidden state.

\begin{figure}[p]
    \centering
    \begin{subfigure}{\linewidth}
        \centering
        \includegraphics[width=\linewidth]{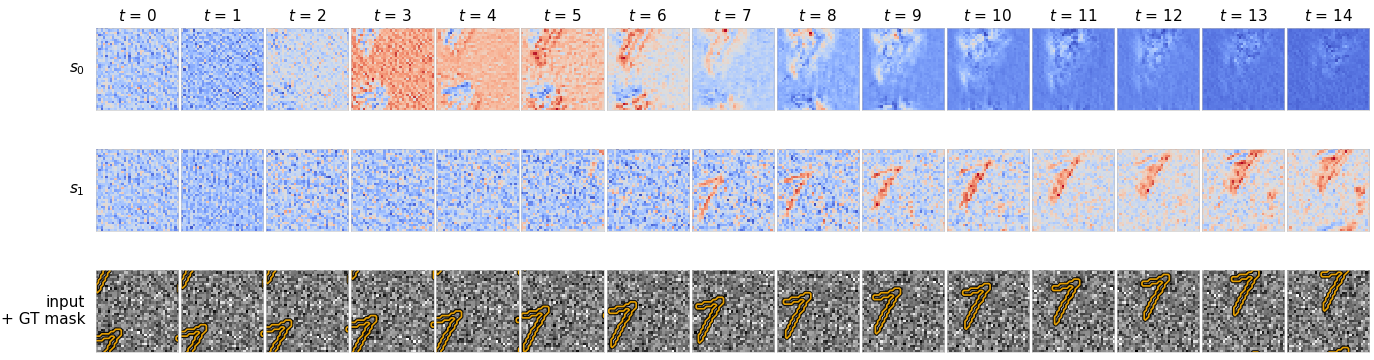}
        \caption{\ME{}}
        \label{fig:cf_states_me}
    \end{subfigure}

    \vspace{0.6em}
    \begin{subfigure}{\linewidth}
        \centering
        \includegraphics[width=\linewidth]{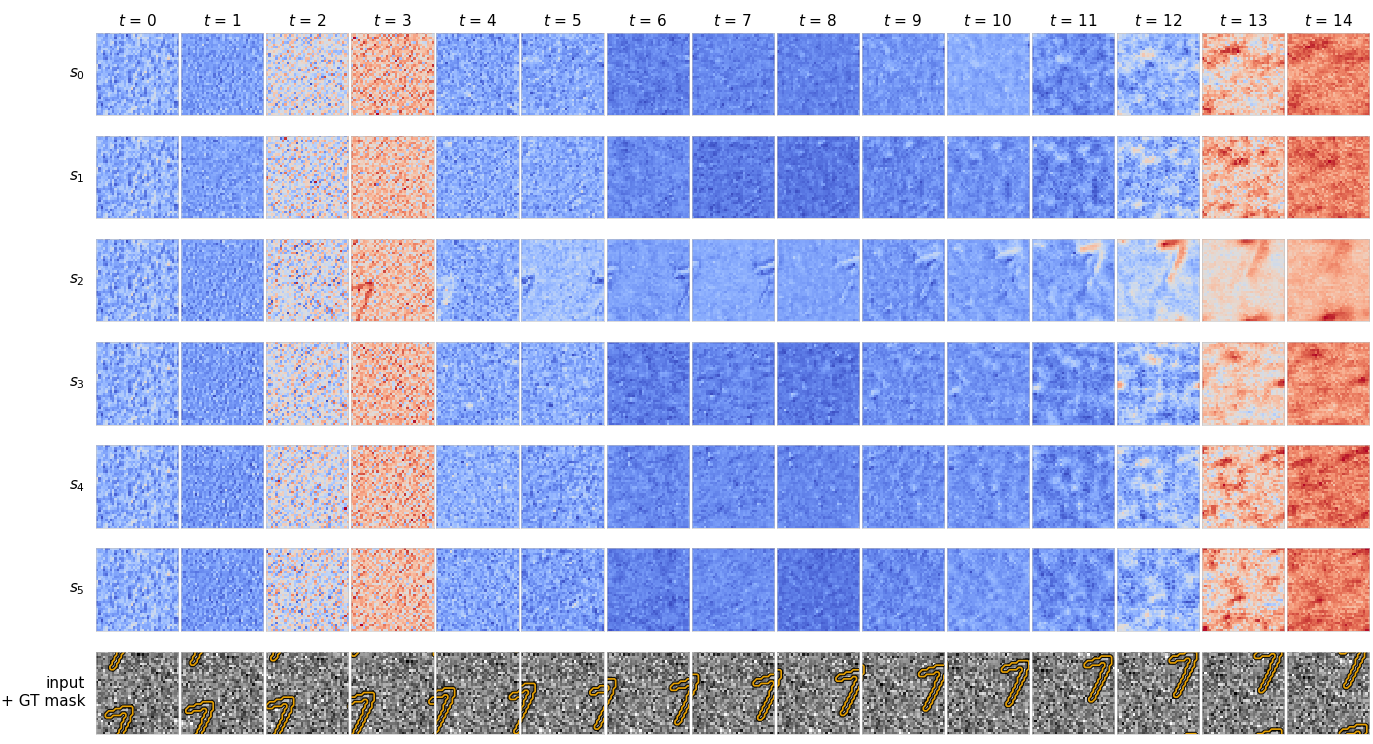}
        \caption{\FE{}}
        \label{fig:cf_states_fe}
    \end{subfigure}
    \caption{\textbf{Hidden states on Motion-Only MNIST.} Channel-averaged state
    of each \ME{} slot (a) and of the \FE{} copies at the velocities the digit
    takes during the context (b); the bottom rows show the input frames. The
    digit's outline is drawn only for the reader: the frames themselves are pure
    noise. \ME{}'s slot $s_1$ follows the digit and accumulates its shape, while
    $s_0$ follows the background; because the digit's velocity keeps changing, no
    \FE{} copy accumulates the shape for long.}
    \label{fig:cf_states}
\end{figure}

\paragraph{Hidden states.}
Figure~\ref{fig:cf_states} shows the hidden states on example sequences: the
digit builds up clearly in the \ME{} slot that follows it, but only faintly in
\FE{}'s copies, since each velocity change moves it to another copy.

\paragraph{Why \CL{} can leave chance.}
Each step of \CL{} sees the current frame within $1$ pixel, through its
$3\times3$ input kernel, and the previous frame within $2$ pixels, through the
hidden state and its $3\times3$ recurrent kernel. One step can therefore compare
the two frames at every displacement up to $3$ pixels, which covers every
velocity in the data, so \CL{} can learn local motion detectors that pick out
the digit. Finding such detectors in noise is hard, however, and \CL{} does so
only on some seeds.

\paragraph{Motion-generalization protocol.}
The sweep is evaluation only: each cell has $5000$ sequences on MNIST test
glyphs, shared by the three models. BatchNorm keeps the statistics stored with
each checkpoint, estimated on the training motion; re-estimating them on each
cell would let the model partly adapt to the new motion. The switching-rate and jump-size axes are those of the future-frame prediction
experiment. That experiment
also orders motion regimes by the unpredictability $\rho$
(Eq.~\eqref{eq:rho}), which compares the one-step error of reusing the previous
velocity with the error of assuming no motion. On white noise every nonzero
shift changes the image by the same amount, so both errors simply count the
frames with a nonzero shift, $\rho$ equals the switching rate, and a $\rho$
panel would repeat the switching-rate panel. For the speed axis, velocities are
drawn from $V_n$ with maximum speed $n\in\{1,3,4,5\}$ instead of
$2$; \FE{}'s lattice contains only the velocities of speed at most $2$. Relative
accuracy divides each model's accuracy in a cell by its accuracy on the training
motion.

\subsection{KTH Action Recognition under Camera Motion}
\label{app:kth}
\paragraph{Data.}
KTH has $599$ videos of $25$ people performing six actions in four scenarios, at
$25$ frames/s and $160\times120$ pixels. Each video is annotated with the segments
in which the action is performed. As in \citet{keller2025flowequivariantrecurrentneural},
the clips are the segments of at least $32$ frames, converted to grayscale and
resized to $32\times32$ without keeping the aspect ratio. People $1$--$16$,
$17$--$20$ and $21$--$25$ give $1{,}517$ training, $380$ validation and $472$ test
clips. The model sees $16$ frames of a clip, taken at every second frame, that is,
at $12.5$ frames/s. In training, this window is drawn at random, and with probability $1/2$ all
its frames are mirrored left to right. In validation and testing, each
clip has one fixed window, not mirrored and shared by all test sets, so the test
sets differ only in the camera motion.

\begin{figure}[tb]
\centering
\includegraphics[width=1\linewidth]{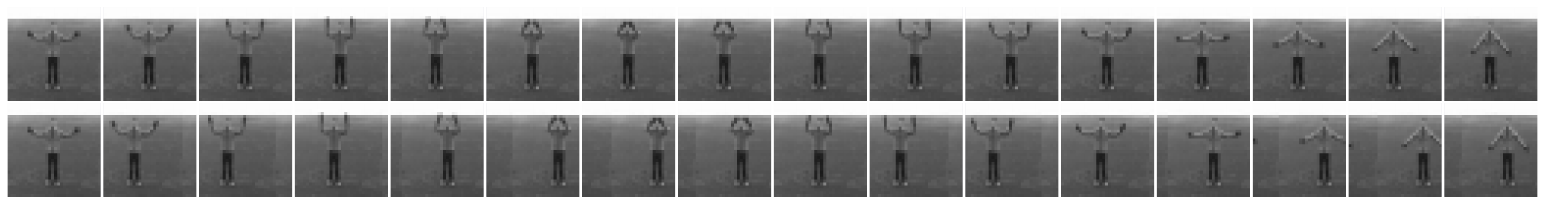}
\caption{\textbf{Example KTH clip with and without camera shake.} All 16 frames of a handwaving clip from the test set. Top: original clip (static camera). Bottom: the same clip with the camera shaking horizontally, with $A = 8$ px and $P = 8$ frames (1.6 Hz). }
\label{fig:kth_example}
\end{figure}

\paragraph{Camera motion.}
The frame at time $t$ is translated by a whole-pixel displacement $d(t)$, with $d(0)=0$ and wrap-around. Since every model uses circular padding, this is an exact group
action without boundary effects. In a shake, each axis of $d(t)$ follows
$A\sin(2\pi t/P+\varphi)$, rounded to whole pixels and shifted so that $d(0)=0$;
the amplitude $A$, the period $P$ in frames and the phase $\varphi$ are drawn once
per clip and per axis, and the frequency is $f=12.5/P$~Hz. The training shake has
$A\in[0,2]$~px and $P\in[16,32]$ ($f=0.4$--$0.8$~Hz). Its fastest step,
$2A\sin(\pi/P)$, stays below $1$~px, so every step lies in $V_1$, and the scene never moves more
than $4$~px from its first position. Amplitudes below $0.5$~px round to no motion,
so $6\%$ of the training clips are static. Validation uses the training shake. See Figure~\ref{fig:kth_example} for an example of a sequence with and without camera shake.

\paragraph{Models.}
All models use the same convolutional encoder, with $64$ hidden channels and $3\times3$
circular kernels ($150{,}016$ parameters), the max over copies or slots of the final
hidden state, and the classifier head of Appendix~\ref{app:cf} with $64$ channels and
an MLP $128\to128\to6$ with leaky ReLU ($128{,}262$ parameters). \FE{} carries $9$
copies at the velocities of $V_1$ while \ME{} has $2$ slots, whose velocities are
estimated to sub-pixel precision and applied with an interpolating warp, and slot
tracking considers every displacement up to half the frame. Its frame-pair
velocities are the two strongest peaks of $\mathcal P_2(\CX_{t-1},\CX_t)$. They
come as an unordered set, so each is given to the slot whose previous velocity is
nearest. A constant camera shifts the new peaks and the previous velocities alike
and leaves this matching unchanged, but a camera whose velocity changes does not,
and a slot can then be given the velocity of another slot's content. Slot tracking
needs no matching, since each slot finds its own content in the next frame, and is
equivariant to any camera path. Frame-pair velocities are therefore used only early
in training: each step $t\ge2$ of each training sequence uses them with a
probability that decays from $1$ to $0$ along a half cosine over the first $200$
epochs, and slot tracking otherwise. Evaluation uses slot tracking only.

\paragraph{Training.}
Every model is trained for $500$ epochs of $48$ batches of $32$ clips, about
$24{,}000$ steps, with Adam, a learning rate decaying from $3\cdot10^{-4}$ to
$10^{-5}$ along a cosine, gradient clipping at $1$ and cross-entropy. Before every
validation and test pass, the head's BatchNorm statistics are re-estimated on $30$
training batches (precise BatchNorm; \citealp{wu2021rethinking}), and the checkpoint
with the best validation accuracy is tested. The seed sets the weights, the data
order, the training windows and the mirroring; evaluation windows and camera paths are
the same for all seeds. On one NVIDIA A100 GPU, the training pass of an epoch
takes $1.3$~s for \CL{}, $9.1$~s for \FE{} and $3.8$~s for \ME{} (median over
epochs).

\paragraph{Shake map.}
Figure~\ref{fig:kth} uses $A\in\{0,1,1.5,2,3,4,6,8\}$~px and
$P\in\{32,24,16,12,8,6,4\}$ frames, with a random phase per axis and clip; $A=0.5$
is left out because it rounds to no motion. The fastest shake moves the camera by
up to $11$~px between frames. A shake stays on \FE{}'s lattice when its fastest
step $2A\sin(\pi/P)$ is at most $1$~px, roughly $A\le P/2\pi$, which holds in $16$
of the moving cells.

\end{document}